\documentclass[conference]{IEEEtran}
\IEEEoverridecommandlockouts
\usepackage[dvipsnames,         svgnames,         x11names]{xcolor}
\newcommand{\eat}[1]{}

\usepackage{xspace}

\newtheorem{example}{Example}

\newcommand{\at}[1]{\protect\ensuremath{\mathsf{#1}}}

\newcommand{\sstab}{\rule{0pt}{8pt}\\[-2.2ex]}
\newcommand{\bi}{\begin{itemize}}
\newcommand{\ei}{\end{itemize}}
\newcommand{\be}{\begin{enumerate}}
\newcommand{\ee}{\end{enumerate}}
\newcommand{\stitle}[1]{\sstab\noindent{\bf #1}}

\newcommand{\ie}{{\em i.e.,}\xspace}
\newcommand{\eg}{{\em e.g.,}\xspace}

\newcommand{\eop}{\hspace*{\fill}\mbox{$\Box$}}     

\newcommand{\term}[1]{{\tt #1}}

\newcommand{\pair}{{\small \sf question}\xspace}

\newcommand{\pairs}{{\small \sf questions}\xspace}

\newcommand{\demo}{{\small \sf demonstration}\xspace}

\newcommand{\demos}{{\small \sf demonstrations}\xspace}

\newcommand{\sys}{\textsc{BatchER}\xspace}

\usepackage{multirow}
\usepackage[normalem]{ulem}
\useunder{\uline}{\ul}{}
\usepackage{cite}
\usepackage{amsmath,amssymb,amsfonts}
\usepackage{algorithm}
\usepackage{algorithmic}
\usepackage{graphicx}
\usepackage{textcomp}
\usepackage{xcolor}
\usepackage{colortbl}
\usepackage{multirow}
\usepackage{booktabs}
\usepackage{balance}
\usepackage{tabularx}
\usepackage{subcaption}
\usepackage{bm}
\usepackage{url}
\usepackage{makecell}

\def\BibTeX{{\rm B\kern-.05em{\sc i\kern-.025em b}\kern-.08em
    T\kern-.1667em\lower.7ex\hbox{E}\kern-.125emX}}
\begin{document}

\title{Cost-Effective In-Context Learning for Entity Resolution: A Design Space Exploration}

\author{\IEEEauthorblockN{
		Meihao Fan\IEEEauthorrefmark{2}, 
		Xiaoyue Han\IEEEauthorrefmark{2},
		Ju Fan\IEEEauthorrefmark{2}, 
		Chengliang Chai\IEEEauthorrefmark{3},
		Nan Tang\IEEEauthorrefmark{4},
		Guoliang Li\IEEEauthorrefmark{1},
		Xiaoyong Du\IEEEauthorrefmark{2}}
	\IEEEauthorblockA{\IEEEauthorrefmark{2}\textit{Renmin University of China}, \IEEEauthorrefmark{3}\textit{Beijing Institute of Technology},\IEEEauthorrefmark{4}\textit{HKUST (GZ)}, \IEEEauthorrefmark{3}\textit{Tsinghua University}\\
		\{fmh1art, cloverhxy, fanj, duyong\}@ruc.edu.cn,
		ccl@bit.edu.cn, nantang@hkust-gz.edu.cn, liguoliang@tsinghua.edu.cn}
}

\maketitle

\begin{abstract}
Entity resolution (ER) is an important data integration task with a wide spectrum of applications. The state-of-the-art solutions on ER rely on pre-trained language models (PLMs), which require fine-tuning on a lot of labeled matching/non-matching entity pairs. 
Recently, large languages models (LLMs), such as GPT-4, have shown the ability to perform many tasks without tuning model parameters, which is known as \emph{in-context learning} (ICL) that facilitates effective learning from a few labeled input context demonstrations. 
However, existing ICL approaches to ER typically necessitate providing a task description and a set of demonstrations for each entity pair and thus have limitations on the monetary cost of interfacing LLMs. 
To address the problem, in this paper, we provide a comprehensive study to investigate how to develop a cost-effective batch prompting approach to ER. We introduce a framework \sys consisting of demonstration selection and question batching and explore different design choices that support batch prompting for ER. We also devise a covering-based demonstration selection strategy that achieves an effective balance between matching accuracy and monetary cost.
%
%
We conduct a thorough evaluation to explore the design space and evaluate our proposed strategies. Through extensive experiments, we find that batch prompting is very cost-effective for ER, compared with not only PLM-based methods fine-tuned with extensive labeled data but also LLM-based methods with manually designed prompting. We also provide guidance for selecting appropriate design choices for batch prompting.
\end{abstract}
\section{Introduction}
\label{sec:intro}

Entity resolution (ER), which finds entities that refer to the same real-world object, is a crucial task for data cleaning and data integration.
Its applications span across various domains, with particular significance in healthcare, finance, customer relationship management, law enforcement, and many others.

The state-of-the-art (SOTA) results in ER are achieved through the application of deep learning methodologies. These methods~\cite{ditto, jointbert, robem, tu2023unicorn, tu2022domain} involve the utilization of Transformer-based models, which are trained on extensive datasets comprising numerous (\eg hundreds or thousands) labeled entity pairs.

\stitle{Standard Prompting and Batch Prompting.}
Meanwhile, large-scale pre-trained language models (LLMs), such as GPT models~\cite{DBLP:conf/nips/BrownMRSKDNSSAA20}, have adopted an emerging learning paradigm called \emph{in-context learning (ICL)}, which does not require to update the model parameters of LLMs~\cite{min2022rethinking, how_many_demo_do_you_need, gao2023ambiguity, wang2023investigating}. It facilitates effective learning from a restricted set of labeled input context demonstrations, referred to as {\bf \demos}.

Next, we use an example to illustrate the typical way of in-context learning, referred to as {\bf standard prompting}.

\begin{figure}[!t]
    \centering 
    \includegraphics[width=0.99\columnwidth]{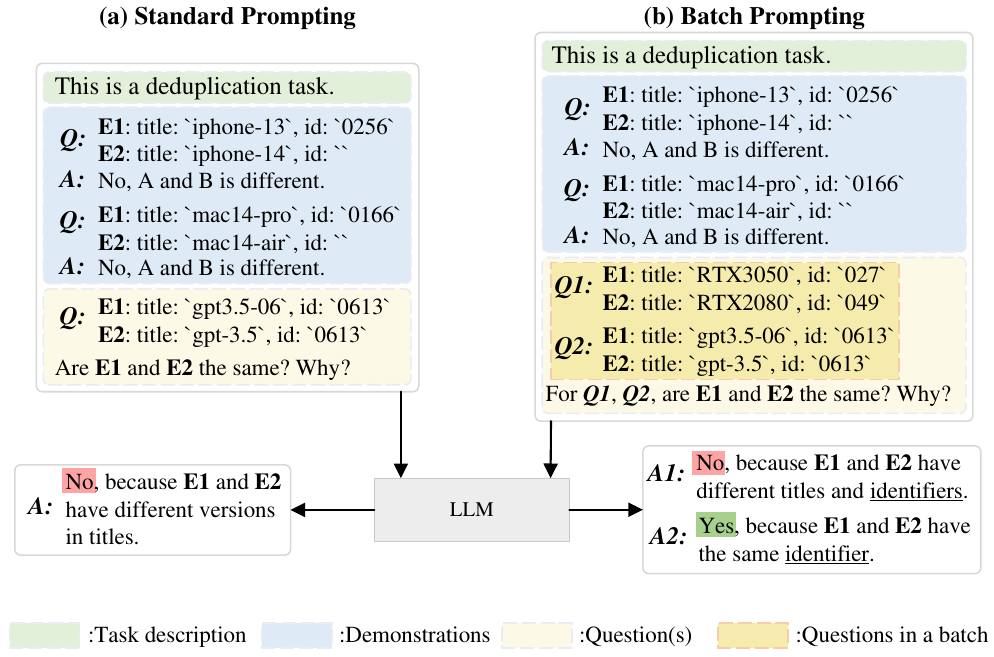}
    \vspace{-1.5em}
    \caption{Standard Prompting and Batch Prompting}
    \label{fig:instance_bp_vs_sp}
    \vspace{-1em}
\end{figure}

\vspace{1ex}
\begin{example}
\label{exam:standard_prompting}
[Standard Prompting]
Figure~\ref{fig:instance_bp_vs_sp}(a) shows standard prompting for ER. The user needs to provide a {\em task description}, several {\em \demos} (\ie the ER pairs with known matching or non-matching labels), and one {\em \pair} (\ie the ER pair whose label is unknown). An LLM (\eg GPT-4) can then answer whether the two entities in the \pair match or not.  \eop
\vspace{1ex}
\end{example}

    
    

Recent studies have shown that standard prompting for ER is effective on \emph{matching accuracy}~\cite{DBLP:journals/pvldb/NarayanCOR22,DBLP:journals/corr/abs-2310-11244}. However, a key limitation of this approach is its \emph{monetary cost} of calling APIs of LLMs, as it necessitates providing a task description and a set of \demos for each \pair, as explained in the following example. For instance, consider a table with 1,000 records that require about 500,000 predictions for ER. Suppose that each pair has $\sim$$60$ words or $\sim$$90$ tokens. Then, querying GPT-4 with standard prompting consisting of $3$ \demos and $1$ \pair will cost $500,000 \times (90\times(3+1)) \times (0.01/1000)=\$1,800$, where the pricing of GPT-4 API services is $\$0.01$ per 1K tokens (\url{https://openai.com/pricing}).

To be cost-effective, a natural alternative is to use a set, or a \emph{batch} of \pairs when prompting the LLMs, which is known as {\bf batch prompting}.

\vspace{1ex}
\begin{example}
\label{exam:batch_prompting}
[Batch Prompting]
As shown in Figure~\ref{fig:instance_bp_vs_sp}(b), the user needs to provide a {\em task description}, a set of \demos, and a set of \pairs. Subsequently, the underlying LLM can answer whether each \pair (\ie entity pair) in this batch matches or not.  \eop
\vspace{1ex}
\end{example}

However, despite some very recent
attempts of batch prompting for general natural language tasks~\cite{DBLP:conf/naacl/RubinHB22, zhang2023makes,li2023unified,agrawal2022context}, as far as we know, exploring the effectiveness of batch prompting for ER under different design choices is not addressed.
To bridge the gaps, we provide a comprehensive study to investigate how to develop a cost-effective batch prompting approach to ER. To achieve this, we introduce a batch prompting framework called \sys that consists of two main modules, demonstration selection and question batching. Based on the framework, we conduct extensive experiments on well-known ER benchmarks to systemically investigate the following two key questions.

\stitle{A Design Space Exploration on Both Accuracy and Cost.}
Due to the importance of ER and the increasing ability of in-context learning, it is highly desired to systemically study batching prompting for ER, under different design choices, on both matching accuracy and monetary cost. To this end, we categorize different choices in question batching and demonstration selection. For question batching, we categorize existing methods as {\em similarity-based}, {\em diversity-based} and {\em random question}. For demonstration selection, we classify existing methods as {\em fixed}, {\em top$k$-\at{batch}}, and {\em top$k$-\at{quesion}}.

\stitle{A Novel Covering-based Selection Strategy.}
While empirically exploring the above design space, we find that existing solutions only consider selecting top-$k$ \demos after a batch of \pairs is determined, without considering whether the selected \demos can well cover all \pairs in a batch. Thus, we further study the problem:``{\em how to select a batch of \pairs and how to select a set of \demos collectively, such that the \demos can well cover all \pairs which can best guide LLMs to provide answers}''?
We model the problem 
as a set cover problem, which is known as NP-hard. 
We solve the problem by devising a covering-based selection strategy, which selects \demos by considering relevance and coverage. The covering-based strategy aims to generate a labeled demonstration set by selecting the minimum number of \demos to cover all \pairs and then labeling them, and thus can effectively balance the trade-off between matching accuracy and monetary cost.

\stitle{A Summary of Experiments.}
We conduct a thorough evaluation to explore the design space and evaluate our proposed strategies. Our experimental findings reveal insights into accuracy and cost of different batch prompting strategies.
(1) Batch prompting can bring 4x-7x cost saving and achieve higher and more stable accuracy than standard prompting.
(2) The design choice that combines diversity-based question batching and our proposed covering-based demonstration selection is the most favorable, \ie achieving the highest accuracy while incurring the lowest cost.
(3) Our \sys framework is the most cost-effective, compared with not only PLM-based methods~\cite{ditto,jointbert,robem} fine-tuned with extensive labeled data, but also LLM-based methods with manually designed prompting~\cite{DBLP:journals/pvldb/NarayanCOR22}.

\stitle{Contributions.}
We make the following notable contributions.

\be
    \item 
    We investigate the design space of batch prompting for ER, by introducing a framework \sys and systematically categorizing existing methods for question batching and demonstration selection in Section~\ref{sec:overview}.

    \item 
    We introduce specific question batching strategies (Section~\ref{sec:sample_batching}) and demonstration selection methods for ER (Section~\ref{sec:demo_sel}). We devise a novel covering-based selection strategy to connect the process of question batching and demonstration selecting in Section~\ref{sec:cover}.

    \item We empirically evaluate our batch prompting framework \sys (Section~\ref{sec:exp}). We make all codes and datasets in our experiments public at Github\footnote{https://github.com/fmh1art/BatchER}. Based on the evaluation, we provide insights on the strengths and limitations of various strategies, which guides designing cost-effective ICL approaches to ER.
   
\ee

\section{Batch Prompting for Entity Resolution: A Design Space Exploration} 
\label{sec:overview} 

\subsection{Entity Resolution} 

Let $T_A$ and $T_B$ be relational tables with $m$ attributes. Each tuple refers to an entity consisting of $m$ properties, \ie for a tuple $a\in T_A$, $a=\{\mathtt{attr}_i, \mathtt{val}_i\}_{i=1}^m$ where $\mathtt{attr}_i$ and $\mathtt{val}_i$ denote the $i$-th attribute name and value respectively. The problem of {\bf entity resolution} (ER) is to identify all the entity pairs $(a,b)\in T_A\times T_B$ that refer to the same object in the real world based on the corresponding attributes. 

An end-to-end ER system consists of a \at{blocker} and a \at{matcher}. The \at{blocker}'s goal is to identify a subset of $T_A\times T_B$ containing candidate pairs with a high probability of being matched~\cite{ditto, papadakis2020blocking, thirumuruganathan2021deep} while the \at{matcher}'s objective is to determine whether each entity pair $(a,b)$ in the above candidate set
refers to the same real-world entity (\ie \at{matching}) or not (\ie \at{non}-\at{matching}). While the design of an effective blocking strategy is beyond the scope of this paper, we employ a widely accepted blocking method~\cite{ditto, thirumuruganathan2021deep, ge2021collaborem} to produce the aforementioned pairwise candidate set. 



\subsection{In-Context Learning} 

{\bf In-context learning (ICL).} It refers to the capability of LLMs to learn from a few \demos in the input context without any parameters updating~\cite{DBLP:conf/nips/BrownMRSKDNSSAA20}. 

{\bf ICL for ER.}
Given any entity pair $(a_i,b_i)$, we utilize a serialization function to serialize it into a text by concatenating all attribute names and values within the entity pair:
\begin{equation}
\label{eq:serialzation_function}
\begin{aligned}
\mathcal{S}((a_i, b_i)) &= \mathcal{S}(a_i) \texttt{[SEP]} \mathcal{S}(b_i)\\
\mathcal{S}(e) &= \mathtt{attr}_1: \mathtt{val}_1 ... \mathtt{attr}_m: \mathtt{val}_m
\end{aligned}
\end{equation} 
where $\texttt{[SEP]}$ is used to separate entities of a pair and $\mathcal{S}$($\cdot$) denotes the serialization function of each data entity $e$.

Then, we construct a prompt consisting of a task description $\mathtt{Desc}$, several serialized pairs with golden labels $\mathtt{Demos}$ (denoted as \demos in this paper) and a serialized pair $\mathtt{Question}$ to be queried (denoted as \pair). By feeding them to an LLM $G$, we generate the target $y$ with the next token prediction, which can be regarded as a conditional text generation problem:
\begin{equation}
\label{eq:in_context_learning}
\begin{aligned}
y= \arg\max_{y\in Y}  P_{G} (y~|\overbrace{~\mathtt{Desc}\oplus \mathtt{Demos}}^{\text{supervision of ER task}} \oplus \mathtt{Question})\\
 \end{aligned}
\end{equation} where $Y=\{$\at{matching}, \at{non}-\at{matching}$\}$ is the label space. 

As Eq.~\ref{eq:in_context_learning} shows, $G$ receives the task’s supervision only from a pre-defined task description ($\mathtt{Desc}$) and the concatenated \demos ($\mathtt{Demos}$). Usually, In-context learning is highly sensitive to the provided \demos and different question selection strategies will bring huge fluctuations in performance~\cite{lu2021fantastically, chen2022relation}. Thus, a comprehensive exploration for selecting beneficial \demos deserves a detailed design.

\subsection{The \sys Framework and Design Space}


\begin{figure}[!t]
   \vspace{-1em}
    \centering 
    \includegraphics[width=0.99\columnwidth]{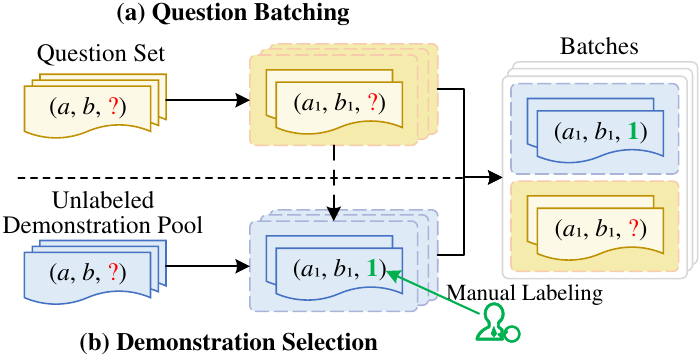}
    \caption{Our proposed \sys framework, which consists of (a) question batching and (b) demonstration selection.}
    \label{fig:batchprompting_for_er}
    \vspace{-0.5em}
\end{figure}

Despite the remarkable accuracy of ICL~\cite{wan2023gptre, agrawal2022context, DBLP:conf/naacl/RubinHB22}, the cost of finance may be very expensive, since most LLM companies such as OpenAI charge users based on the token consumption.

To reduce the cost of interfacing LLMs while maintaining high accuracy, batch prompting is proposed, which allows to query a batch of \pairs with several \demos and asks LLM to make multiple predictions in one interface~\cite{cheng2023batch}. 

\vspace{1ex}
\begin{example}
Figure~\ref{fig:instance_bp_vs_sp} shows the difference between Standard Prompting and Batch Prompting. Although both select two \demos for LLMs to lean in context, Batch Prompting asks LLMs to answer 2 \pairs at one interface, which approximately saves tokens of $2$ \demos and $1$ task descriptions. Naturally, the more \pairs we put in a batch, the more cost of interfacing LLMs will be reduced.
\eop
\vspace{1ex}
\end{example}

\stitle{The \sys Framework.}
We can observe that two critical components in the prompt of Batch Prompting are in-context \demos and \pairs. Thus, to design effective Batch Prompting, we introduce a framework called \sys that consists of the modules of in-context demonstration selection and question batching, as shown in Figure~\ref{fig:batchprompting_for_er}. The \sys framework takes a set of \pairs, \ie entity pairs $\{q\}$ as input, and aims to produce a set of \emph{batch prompts}, which are then fed into an LLM. As a prompt needs in-context \demos, \sys also considers a set of entity pairs without \at{matching}/\at{non}-\at{matching} results as an {Unlabeled Demonstration Pool}.
In this section, we first formally define the above two modules and then systematically explore the design space of Batch Prompting for ER by categorizing each individual module in the \sys framework. 
%
\bi
\item 
\textbf{Question Batching.} Considering a Question Set $M$ of \pairs to be queried, Question Batching aims to iteratively select $b$ \pairs and group them into one batch $B_i=\{q_j\}_{j=1}^b$. To ensure all \pairs will be queried at least once, the union set of all batches should equal to the original question set, satisfying $\bigcup B_i = M$.
\item 
\textbf{Demonstration Selection.} Considering a large pool of unlabeled \demos $D_u$ from which we iteratively select several data points $\{d_j\}$ for each batch $B_i$. We assume manual annotation will be adopted for the selected data to generate labeled \demos $D_i=\{(d_j, y)\}$ which will be used to guide LLMs to make predictions for batched \pairs. 
\ei

To put the above together, the \sys Framework takes a Question Set $M$ and an Unlabeled Demonstration Pool $D_u$ as input and outputs a set of question batches $B=\{B_i\}$ along with a set of corresponding \demos $D=\{D_i\}$, satisfying $\bigcup B_i=M$ and $\bigcup D_i\subseteq D_u$.

\begin{table}[t!]
\vspace{-1em}
\centering
\caption{\bf{A Design Space Exploration}}
\label{tbl:design_space}
\vspace{-1ex}
\renewcommand{\arraystretch}{1.12}
\begin{tabular}{|cl|}
\hline
\multicolumn{1}{|c|}{\textbf{Modules}}                            & \multicolumn{1}{c|}{\textbf{Categorization}} \\ \hline \hline
\multicolumn{1}{|c|}{\multirow{3}{*}{\makecell[c]{Question Batching}}} & (1) Random                 \\ \cline{2-2} 
\multicolumn{1}{|c|}{}                                   & (2) Similarity-based                    \\ \cline{2-2} 
\multicolumn{1}{|c|}{}                                   & (3) Diversity-based         \\ \hline \hline
\multicolumn{1}{|c|}{\multirow{4}{*}{\makecell[c]{Demonstration Selection} }} & (1) Fixed                     \\ \cline{2-2} 
\multicolumn{1}{|c|}{}                                   & (2) Top$k$-\at{batch}              \\ \cline{2-2} 
\multicolumn{1}{|c|}{}                                   & (3) Top$k$-\at{question} 
\\ \cline{2-2} 
\multicolumn{1}{|c|}{}                                   & \makecell[c]{\textbf{(4) Covering-based}  {\bf (Our proposal)} }        \\ \hline
\end{tabular}
\vspace{-1em}
\end{table}

\stitle{A Design Space Exploration.}
To utilize in-context learning for ER, several challenges should be addressed. First, the question batching and demonstration selection require a feature extractor to map \pairs and \demos into a vector space, which facilitates the measurement of their relevance. However, the widely used semantics-based feature extractor may fail to select beneficial \demos due to the lack of task-specific signals~\cite{wan2023gptre}. Second, although in-context learning shows stable and remarkable performance in Standard Prompting with relevant demonstration selection~\cite{luo2023dr,margatina2023active, agrawal2022context}, effective demonstration selection strategies still lacks a comprehensive investigation on the trade-off between accuracy and cost. At last, 
the choice of batching strategy is of great significance in downstream performance, which deserves in-depth investigation.

To address the challenges, we propose a categorization of design choices for each module in \sys, which forms a design space as shown in Table~\ref{tbl:design_space}. 
We first explore strategies for the question batching module and discuss different feature extractors used for measuring relevance among \pairs (Section~\ref{sec:sample_batching}). Subsequently, we investigate methods for selecting \demos for a batch (Section~\ref{sec:demo_sel}). We note that \sys is extensible, \ie it is possible
to incorporate new modules, new categories, or new methods or variants of existing methods. Moreover, it is possible to define the search space from a different angle; that is, we contend that our proposal is rational, but may not be unique.

%

\section{Question Batching}
\label{sec:sample_batching}

\begin{figure}[!t]
    \centering 
    \includegraphics[width=0.99\columnwidth]{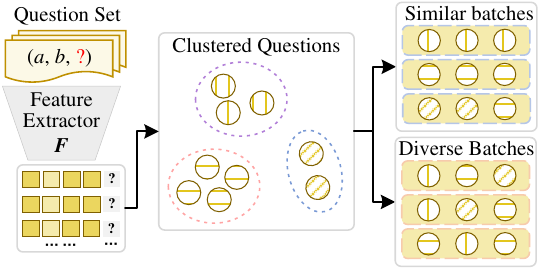}
    \vspace{-1.5em}
    \caption{Question Batching Framework}
    \label{fig:batching_framework}
    \vspace{-0.5em}
\end{figure}

This section explores the question batching strategies, as shown in Table~\ref{tbl:design_space}. To this end, we first describe a general framework of question batching, as illustrated in Figure~\ref{fig:batching_framework}.
Specifically, given a Question Set $M$ of entity pairs, the framework produces batches of \pairs in three steps. 
\bi
    \item \textbf{Feature Extraction.} We first use a Feature Extractor to cast the \pairs into feature vectors
    \item \textbf{Question Clustering.} We then adopt an unsupervised clustering algorithm such as DBSCAN or K-Means to group the \pairs into clusters.
    \item \textbf{Question Batching.} We finally group \pairs into batches based on the clusters using various strategies.
\ei 


In the remaining of this section, we mainly introduce three representative batching strategies, including random question batching, similarity-based question batching, and diversity-based question batching, which have been adopted by previous studies~\cite{cheng2023batch, zhang2023large} (Section~\ref{subsec:batch-strategies}).
Next, as feature extraction and distance measurement (for clustering) are involved in the batching process, we then discuss two feature extraction methods
in Section~\ref{subsec:feature_extractor}.
Note that, for question clustering, we adopt DBSCAN~\cite{DBLP:conf/kdd/EsterKSX96}, as the algorithm achieves the best performance. Due to the space limit, this section does not discuss various clustering algorithms, which are not the focus of this paper.

\subsection{Batching Strategies} 
\label{subsec:batch-strategies}

Given clustered \pairs, \sys generates batches based on the following three representative strategies, random question batching, similarity-based question batching, and diversity-based question batching, which have been adopted by previous studies~\cite{cheng2023batch, zhang2023large}.

\stitle{Similarity-based Question Batching.} The intuition of this strategy is to group \emph{similar} \pairs within the same clusters into the same batch. To this end, we iteratively select $b$ (\ie batch size) \pairs from the same cluster to form a batch, to ensure that \pairs in the same batch have similar feature vectors to each other.
In particular, during the final stage of batch generation, some clusters may contain \pairs fewer than the required batch size $b$. 
%
%
In such case, 
we select the largest remaining cluster, denoted as $C_\text{max}$. We then seek to pair it with another cluster whose size exactly matches $b - |C_\text{max}|$, to form a complete batch. If no such cluster exists, we opt for the next largest cluster, randomly selecting $b - |C_\text{max}|$ elements from them to form a batch in conjunction with $C_\text{max}$. 


\stitle{Diversity-based Question Batching.} The intuition of this strategy is to group \pairs that are from diversified clusters into a batch. In this batching strategy,
batches are also generated in two stages. Firstly, we ensure batch diversity by selecting one \pair from each of $b$ different clusters, such that the \pairs in different batches have obvious differences in feature vectors from each other. Then, when the batching process nears completion, we may encounter scenarios where the number of available clusters is less than $b$. In such instance, we simply ensure the diversity of batches generated from a limited number of clusters by selecting \pairs from remaining clusters in a round-robin manner. 

\vspace{1ex}
\begin{example}
\label{exam:question_batching}
[Question Batching]
Consider the \pairs in Figure~\ref{fig:batching_framework}. We denote the three clusters as $C_a=\{q^a_i\}_{i=1}^2$, $C_b=\{q^b_i\}_{i=1}^3$, and $C_c=\{q^c_i\}_{i=1}^4$, respectively. 

(1) For similarity-based question batching, we sequentially select $C_b$ and $C_c$, forming batches $B_1=\{q^b_1, q^b_2, q^b_3\}$ and $B_2=\{q^c_1, q^c_2, q^c_3\}$. Subsequently, from the remaining clusters $C_a=\{q^a_1, q^a_2\}$ and $C_c=\{q^c_4\}$, we choose the larger cluster $C_a$ and combine it with $C_c$ to create $B_3=\{q^a_1, q^a_2, q^c_4\}$. 

(2) For diversity-based question batching, we can generate diverse batches $B_1=\{q^a_1, q^b_1, q^c_1\}$ and $B_2=\{q^a_2, q^b_2, q^c_2\}$ in the initial stages by iteratively selecting one \pair from $C_a$, $C_b$ and $C_c$. Then with remaining clusters $C_b=\{q^b_3\}$ and $C_c=\{q^c_3, q^c_4\}$, we sequentially select \pairs from $C_c$, $C_b$ and $C_c$ to generate the final batch $B_3=\{q^c_3, q^b_3, q^c_4\}$. \eop
\vspace{1ex}
\end{example}
%

\stitle{Random Question Batching.} We also consider a straightforward random question batching strategy, which is commonly adopted in the existing works~\cite{cheng2023batch, zhang2023large}.
%
In this approach, each batch is formed by randomly selecting \pairs from the remaining question set. Due to this randomness, the generated batches may contain a mix of both similar and dissimilar \pairs. This implies that a random batch, to some extent, represents a middle ground between a similar batch and a diverse batch. 

\subsection{Feature Extractor}
\label{subsec:feature_extractor} 

The process of batching \pairs in the previous section relies on the utilization of a feature extractor to convert \pairs into corresponding feature vectors. Subsequently, these feature vectors are used to calculate distances between \pairs and then serve as the basis for the clustering procedure. Formally, given a set of \pairs $M$, we need to define a feature extractor $f$ and a distance function $\mathtt{dist}$, and thus the distance of any two \pairs $q_i$ and $q_j$ can be calculated via $\mathtt{dist}(\mathbf{v}_i,\mathbf{v}_j)$ between the two feature vectors, \ie $\mathbf{v}_i$ and $\mathbf{v}_j$.
%
%
We notice that the distance function can be further defined by a variety of ways, such as Euclidean distance or cosine similarity (distance). In our experiments, we define the distance function based on the Euclidean distance, which achieves the best performance among others. 

Next, we introduce two types of feature extractors, one based on semantics and the other being structure-aware.


\stitle{Semantics-based Feature Extractor.} Semantics-based feature extractor utilizes a pre-trained language model (PLM) to encode each serialized \pair. For ER task, as all \pairs are structural pairs, \ie with multiple attributes, we first use the serialization function defined in Eq.(\ref{eq:serialzation_function}) to generate serialized \pairs and pass it to a PLM, such as SBERT~\cite{reimers2019sentence} and RoBerta~\cite{liu2019roberta} to generate embedding-based representations. Formally, given a \pair $q$, the feature vector $\mathbf{v}$ can be generated as follow:
\begin{equation}
\label{eq:emb_plm} 
\begin{aligned}
\mathbf{v} = \texttt{Encoder}(\mathcal{S}(q))
\end{aligned}
\end{equation} where $\texttt{Encoder}$ denotes the encoding function of a PLM.
Although the above feature extractor formulates the relevance as semantic distance, it may have the limitation of ignoring the structural information. This inspires us to introduce another feature extraction method, which can capture structural similarity to model relevance. 

\stitle{Structure-aware Feature Extractor.} Structure-aware feature extractor employs a string similarity function to map attribute-matching signals of two entities of a \pair into a low-dimensional space, which enables the generated feature vectors to capture structural information and task-related knowledge. Formally, given a structural pair $(a, b)$, we derive the feature vector by calculating the similarities of attributes between $a$ and $b$. Since attribute values typically take a string format, we can compute similarity $s_i$ on attribute $\mathtt{attr}_i$ with string similarity function, e.g., Levenshtein ratio and Jaccard. 

Using the Jaccard similarity, we tokenize $\mathtt{val}^a_i$ and $\mathtt{val}^b_i$ into sets and compute the similarity as:
\begin{equation}
\label{eq:sim_jac}
s_i = \mathtt{JAC}(\mathtt{val}^a_i, \mathtt{val}^b_i) = \frac{|\mathtt{val}^a_i\cap \mathtt{val}^b_i|}{|\mathtt{val}^a_i\cup \mathtt{val}^b_i|}
\end{equation} where $\mathtt{val}^a_i$ represents the tokenized set of attribute value $\mathtt{val}_i$ of entity $a$ and $|\mathtt{val}^a_i|$ represents corresponding token-set size.

The Levenshtein ration (LR) derives from the Levenshtein edit distance (LED)~\cite{levenshtein1966binary}, representing the minimum number of edits needed to transform one string into another, as: 
%
\begin{equation}
\label{eq:sim_lr}
s_i = \mathtt{LR}(\mathtt{val}^a_i, \mathtt{val}^b_i) = 1-\frac{\mathtt{LED}(\mathtt{val}^a_i, \mathtt{val}^b_i)}{s}
\end{equation} where $\mathtt{LED}$ is the Levenshtein edit distance function and $s$ represents the sum of string length of $\mathtt{val}^a_i$ and $\mathtt{val}^b_i$.

Thus, given a \pair $q$ with entity pair $(a, b)$, the feature vector $\mathbf{v}$ can be generated by concatenating the similarities of all attributes make $\mathbf{v}=\{s_i\}_{i=1}^m$.

\begin{figure}[!t]
\vspace{-1em}
    \centering 
    \includegraphics[width=0.99\columnwidth]{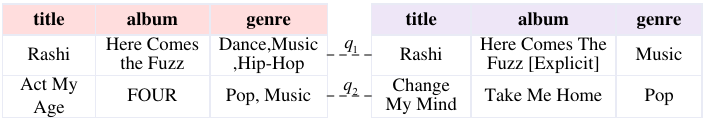}
    \caption{An example instance of Entity Resolution.}
    \label{fig:instance_entity_resolution}
    \vspace{-1em}
\end{figure}

\vspace{1ex}
\begin{example}
\label{exam:feature_extractor}
[Feature Extraction]
Figure~\ref{fig:instance_entity_resolution} shows an example instance of entity resolution. 

(1) For semantics-based feature extractor, we first serialize $q_1$ with Eq.~\ref{eq:serialzation_function} and obtain $S(q_1)=$“title:Rashi, album:Here..., genre:Dance... $\texttt{[SEP]}$ title:Rashi, album:Here..., genre:Music”. Then we utilize a pre-trained language model such as SBERT to encode the embedding as feature vector $\mathbf{v}_1$. 

(2) For structure-aware feature extractor, to generate $\mathbf{v}_1$ for $q_1$, we first compute the string similarities of “Rashi” and “Rashi”, “Here Comes the Fuzz” and “Here Comes The Fuzz [Explicit]”,  and “Dance,Music,Hip-Hop” and “Music”. Suppose we utilize $\mathtt{LR}$ function, the similarities of title, album, and genre can be computed as 1, 0.73, and 0.42. Second, the similarities are concatenated to make up the feature vector $\mathbf{v}_1=[1,0.73,0.42]$. Similarly, the feature vector of $q_2$ can be computed as $\mathbf{v}_2=[0.33,0,0.46]$. 
%
%
\eop
\vspace{1ex}
\end{example}

\section{Demonstration Selection}
\label{sec:demo_sel}

\begin{figure}[!t]
   \vspace{-1em}
    \centering 
    \includegraphics[width=0.99\columnwidth]{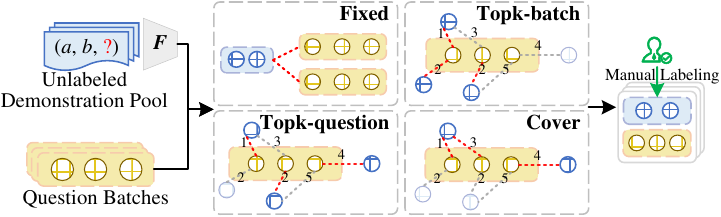}
    \caption{Demonstration Selection Framework, where blue circles and yellow circles represent \demos and \pairs respectively, and values on edges represent distances. }
    \label{fig:example_selection_framework}
    \vspace{-1em}
\end{figure}

 Figure~\ref{fig:example_selection_framework} illustrates the framework of demonstration selection and describes four demonstration selection methods. Given an Unlabeled Demonstration Pool $D_u$ and a set of generated question batches $B$, demonstration selection aims to select beneficial in-context \demos $D_i$ for each batch $B_i\in B$, which will be then manually labeled. To further specify the concept of four demonstration selection methods, we give an illustration for each method. For simplicity, we only consider two closest \demos for each \pair.

\subsection{Fixed Demonstration Selection} 

A basic idea is to sample fixed $K$ \demos and then allocate them to every batch. In Figure~\ref{fig:example_selection_framework}, we generate two fixed \demos by randomly sampling from the unlabeled demonstration pool and allocate these two \demos to each batch. This method brings a fixed annotation cost. However, existing studies show that random \demos may incur unstable performance of ICL~\cite{lu2021fantastically, chen2022relation}. 

\subsection{Top$k$-\at{batch} Demonstration Selection} 

Similar to the strategy in Standard Prompting of recommending top $k$ most relevant \demos~\cite{DBLP:conf/acl-deelio/LiuSZDCC22}, this strategy selects the $k$ most relevant \demos for each batch. Since a batch $B_i$ and a \demo $d$ are not in the same dimension, we first define the relevance between $B_i$ and $d$ based on the distance function $\mathtt{dist}$ defined in Section~\ref{subsec:feature_extractor}:
\begin{equation}
\label{eq:rel_between_batch_demo}
\mathtt{dist}^*(B_i, d)=\min_{q_j\in B_i} \mathtt{dist}(q_j, d)
\end{equation}
which shows that we define the relevance between $B_i$ and $d$ as the minimum distance between $d$ and all \pairs in the batch. Based on this, we can use the $k\mathtt{NN}$ algorithm to generate $k$ in-context \demos for $B_i$ by $D_i=k\mathtt{NN}(B_i, D_u)$. 
In Figure~\ref{fig:example_selection_framework}, we set $k$ as batch size $|B_i|$, and thus Top$k$-\at{batch} sequentially selects $k$ \demos (bold blue circles) based on the $k$ shortest edges (red dotted lines).

However, this method may not be able to assign relevant \demos for some particular \pairs in a batch. Thus, for such \pairs, the LLM may fail in finding relevant \demos for reference to provide the correct answers. 

\subsection{Top$k$-\at{question} Demonstration Selection} 

To address the above issue, we investigate a demonstration selection method that select the $k$ most relevant \demos for each \pair in the batch. This is based on the assumption that, since relevant \demos are beneficial when querying the individual \pair, the set of relevant \demos will also benefit when querying the whole batch. Formally, considering a batch $B_i=\{q_i\}_{i=1}^b$, the in-context demonstration set $D_i$ can be generated as $D_i=\bigcup_{q_j\in B_i} k\mathtt{NN}(q_j, D_u)$. Figure~\ref{fig:example_selection_framework} illustrates the basic idea of the Top$k$-\at{question} method where we set $k=1$ and select the most relevant \demo for each \pair in the batch.

Although this method is likely to improve the accuracy of ICL, it may have a limitation of incurring large monetary cost. Also, it may generate long prompts which could lead to long text comprehension issue and input length overrun. 


\subsection{Covering-based Demonstration Selection} 
A key limitation of Top$k$-\at{question} and Top$k$-\at{batch} is that they may incur substantial labeling cost, which is caused by labeling the selected \demos. 
%
To mitigate this, we introduce a new approach based on the idea of using \demos to \textbf{“cover”} all \pairs in the batch $B_i$ where “cover” means that the distance between \pair $p$ and \demo $d$ is smaller than a threshold $t$. This is based on the assumption that the beneficial \demos are a set of relevant data points and all beneficial to a given \pair. In Figure~\ref{fig:example_selection_framework}, we assume that \demos with a shorter distance than 5 can be regarded as a beneficial reference when answering the \pair. Thus, we first select the top \demo to cover the left two \pairs. Then, to cover the last \pair, the rightest \demo is selected. 

It is important to recognize that for the given batches, multiple selection choices that fulfill the aforementioned covering-based criteria exist. Thus, in Section~\ref{sec:cover}, we will formally formulate this the covering-based problem and propose an efficient algorithm to solve the problem.

\section{Covering-based Demonstration Selection}
\label{sec:cover} 

The covering-based method aims to address two main problems. First, we need to select a minimal subset of \demos from an unlabeled demonstration pool to cover all the \pairs of all batches. Then, for each batch, we need to further select some \demos from this subset, ensuring the covering of each \pair in the batch while minimizing the total number of tokens. Below, we name these two problems as the Demonstration Set Generation and Batch Covering problems and provide their detailed definitions.

\subsection{Demonstration Set Generation}
\label{sec:demo_set_generation}

\stitle{Definition.} Given a Question Set $M$ containing all \pairs to be queried, an unlabeled demonstration pool $D_u$ and a non-negative distance threshold $t$, we need to select a subset of \demos $D_s \subset D_u$, satisfying $\forall q\in M$, exists at least one $d\in D_s, \mathtt{dist}(q, d) < t$. The goal is to minimize the size of selected Demonstration Set $|D_s|$. 

\stitle{NP-hard Proof Sketch.} 
We can prove the Demonstration Set Generation Problem to be NP-hard by a reduction from the Set Cover Problem, which is proven to be NP-hard~\cite{bernhard2008combinatorial}. 

An instance of Set Cover Problem (SCP) encompasses a universe of items $U$, a collection $V=\{S_1, S_2, S_3,..., S_m\}$ of subsets of $U$, we need to find a subset-collection $V^*\subset V$ such that each element in $U$ is covered by at least one subset in $V^*$. The goal is to minimize the number of selected subsets $|V^*|$.

We reduce SCP to our problem. We show that for any instance $(U,V)$ of SCP, we can create a corresponding instance of our problem based on $(U,V)$ in polynomial time. First, We translate the set $U$ of universal items into the set $M$ of \pairs. Then, Given items $u_j\in U$, if $u_j\in S_i$, we add a demonstration $d_j$ to the unlabeled demonstration pool $D_u$ and set the distance between $d_j$ and $q_i$ to be $0$ (less than $t$). Finally, given the above reduction, we can deduce that the objective of finding the minimum number of subsets in $V$ that cover all items in $U$ in SCP is equivalent to the objective of our problem, which is to find the minimum number of \demos in $D_u$ that cover all \pairs in $M$. 




\begin{algorithm}[t!]
    \caption{Demonstration Set Generation/Batch Covering}\label{alg:covering_based_algorithm}
    {\small
    \begin{algorithmic}[1]
    \renewcommand{\algorithmicrequire}{ \textbf{Input:}}
    \REQUIRE Set of \pairs $Q$, set of \demos $D$, nondecreasing value function $f$, weight function $w$.
    \renewcommand{\algorithmicrequire}{ \textbf{Output:}}
    \REQUIRE set of selected \demos $D_s$.

    \STATE $D_s \gets \varnothing$

    \WHILE{$f_Q(D_s) \neq f_Q(D)$}
        \STATE $d \gets \mathop{\arg\max}\limits_{d\in D} \frac{f_Q(D_s \cup \{d\})-f_Q(D_s)}{w(d)}$
        \STATE $D_s \gets D_s \cup \{d\}$
    \ENDWHILE
    \end{algorithmic}
    }
\end{algorithm}

\stitle{Greedy Algorithm.} To efficiently address the Demonstration Set Generation Problem, we propose a greedy-based algorithm. To start with, we define a non-decreasing value function $f_Q(D_s) = \sum_{i=1}^{|M|} z_i$ to measure the value of intermediate demonstration set $D_s$, where for $q_i\in Q$, $z_i=1$ if $\min_{d_j\in D_s}\mathtt{dist}(q_i, d_j)<t$, otherwise, $z_i=0$. Generally, the value function calculates the number of covered \pairs by $D_s$. Then, taking the value function $f$, set of \pairs $M$, and an unlabeled demonstration set $D_u$ as input, we iteratively select the most efficient \demo. Efficiency is defined by the ratio of the incremental value a \demo contributes to the intermediate Demonstration Set $D_s$ relative to its weight. For the Demonstration Set Generation Problem, we set the weights of all \demos to be 1, since selecting any \demo brings us equivalent cost. The pseudo-code is shown in Algorithm~\ref{alg:covering_based_algorithm}.

We first initialize the demonstration set $D_s$ to an empty set (line 1). Then we determine whether the value of intermediate set $D_s$ meets the value of full unlabeled demonstration pool $D_u$ (line 2) which is probably equaled to $|M|$ with a large enough pool size. If not, we will iteratively select the most efficient \demo and add it to the intermediate demonstration set (lines 3$\sim$4). Otherwise, the algorithm ends and outputs the selected demonstration set $D_s$ (line 5). 

Assuming that the optimal sum of Demonstration Set Generation Problem is $OPT$ and the final sum of our greedy algorithm is $ans^*$, we have $ans^*\leq H_k\cdot OPT$, where $H_k=\sum_{i=1}^k \frac{1}{i}, k=\max_{d_i\in D_s} f_Q(\{d_i\})$. A complete proof can be found in~\cite{slavik1996tight}.

For Demonstration Set Generation problem, by setting a target function and designing a greedy algorithm to optimize it, we can generate an effective solution, that is, selecting a small number of \demos to cover all the \pairs to be queried, thereby greatly reducing the labeling cost. 

\begin{table}[t!]
\centering
\caption{\bf{Statistics of Datasets.}} 
\renewcommand{\arraystretch}{1.12}
\tabcolsep=0.17cm
\vspace{-1mm}
\begin{tabular}{|c|c|c|c|c|}
\hline
\textbf{Dataset}  & \textbf{Domain}     & \textbf{\# Attr.} & \textbf{\# Pairs} & \textbf{\# Matches} \\ 
\hline \hline
Walmart-Amazon (WA)    & Electronics    & $5$       & $10,242$   & $962$        \\ \hline
Abt-Buy (AB)  & Product    & $3$      & $9575$   & $1028$        \\\hline
Amazon-Google (AG)       & Software   & $3$       & $11,460$   & $1,167$      \\\hline
DBLP-Scholar (DS)       & Citation   & $4$       & $28,707$   & $5,347$      \\\hline
DBLP-ACM (DA) & Citation   & $4$       & $12,363$   & $2,220$      \\\hline
Fodors-Zagats (FZ)    & Restaurant & $6$       & $946$      & $110$        \\\hline
iTunes-Amazon (IA)   & Music      & $8$       & $532$      & $132$        \\\hline
Beer     & Beer       & $4$       & $450$      & $68$         \\\hline
\end{tabular}
\label{tbl:datasets}
\vspace{-2em}
\end{table}

\subsection{Batch Covering}

Next, based on the generated Demonstration Set, we will allocate relevant \demos to each batch, so as to covering all the \pairs in the batch. At this stage, we ask a question: Is there further optimization space when allocating \demos? To answer this question, we consider an example of a Question Set $M=\{q_1,q_2,q_3,q_4\}$ and a labeled Demonstration Set $\{d_1,d_2\}$. We have $d_1$ covers $q_1,q_2,q_3$ and $d_2$ covers $q_2,q_3,q_4$. Given a batch $B_i=\{q_2,q_3\}$, we need to allocate \demos to cover all \pairs in $B_i$. It can be seen that, at this time, whether allocating $d_1$ or $d_2$ can cover all questions in the batch. Therefore, although we only consider covering each question once when generating the Demonstration Set, there is still room for choice when allocating \demos for each batch.

\stitle{Definition.} Given a batch $B_i$ of \pairs $B_i =\{q\}$, a generated demonstration set $D_s$ and a non-negative distance threshold $t$, we need to select a set of \demos $D_i \subset D$, satisfying $\forall q\in B_i$, exists at least one $d\in D_i$ such that $\mathtt{dist}(q, d)<t$. The goal is to minimize the weight of selected demonstrations $\sum_{d\in D_i}w(d)$. 

We define the weights of \demos as token numbers, and the goal of our problem is to find a demonstration set to cover the batch with minimum token assumption. 

\stitle{NP-hard Proof Sketch.} The batch covering problem is obviously a special case of the set cover problem when we set the weight of all \demos to be 1. 
Following the proof in section~\ref{sec:demo_set_generation}, we can create a corresponding instance of batch covering problem based on any instance of SCP. Besides, since we set all the weights to be 1, the objective of our problem becomes $\sum_{d\in D_i}1=|D_i|$, which is equivalent to that of SCP. Thus, we can prove the batch covering problem as an NP-hard problem by reducing it from the NP-hard set cover problem.

\stitle{Greedy Algorithm.} We again use Algorithm~\ref{alg:covering_based_algorithm} to address the Batch Covering Problem. We use the same value function defined in section~\ref{sec:demo_set_generation} and define the weights of \demos as token numbers. Taking the value function $f$, batch $B_i$ of \pairs, the generated Demonstration Set $D_s$, and weight function $w$ as input, the algorithm will output the allocated demonstration set $D_i$ for batch $B_i$. 

This greedy algorithm yields an approximation ratio of $\ln |B_i|-\ln\ln |B_i|+\Omega(1)$. A complete proof can be found in ~\cite{slavik1996tight}. 

For Batch Covering Problem, by defining the weights of \demos as token numbers and formulating it as Weighted Set Cover Problem, we can generate an effective solution with the minimum sum of tokens of batch prompts, thereby reducing the interfacing API cost. 

\section{Experiments}
\label{sec:exp}
This section evaluates our batch prompting framework \sys investigated in this paper.
Specifically, we first present the experimental setup in Section~\ref{subsec:exp-setup}, and then conduct experiments to answer the following key questions:

\stitle{Exp-1:} How does \emph{Batch Prompting} compare with \emph{Standard Prompting}? (Section~\ref{subsec:exp-batch-vs-standard})

\stitle{Exp-2:} What are effective strategies in our design space of question batching and demonstration selection? (Section~\ref{subsec:exp-design-space})

\stitle{Exp-3:} How does our proposed \sys framework compare with PLM-based approaches to ER? (Section~\ref{subsec:exp-compare-plms})

\stitle{Exp-4:} How does our proposed \sys framework compare with LLM-based approaches to ER? (Section~\ref{subsec:exp-compare-llms})

\stitle{Exp-5:} What is performance of our \sys framework given various underlying LLMs? (Section~\ref{subsec:exp-given-llms})

\stitle{Exp-6:} What is performance of our \sys framework given different feature extractors? (Section~\ref{subsec:exp-given-feat-extract})

\subsection{Experimental Setup} \label{subsec:exp-setup}

\stitle{Datasets.} 
We evaluate our proposed batch prompting framework \sys using well-adopted benchmarking datasets from Magellan benchmark~\cite{DBLP:journals/cacm/DoanKCGPCMC20}, which range from a variety of domains, such as product, software, and citation.
Table~\ref{tbl:datasets} provides detailed statistics of the datasets.
Specifically, each dataset contains entities from two relational tables with multiple attributes, and a set of labeled matching/non-matching entity pairs. Take the Amazon-Google (AG) dataset as an example: it contains software products from Amazon and Google with three attributes (\term{title}, \term{manufacturer}, \term{price}), and has $11,460$ entity pairs where $1,167$ pairs are matches.
For fair comparison, the set of labeled entity pairs is split into train, validation and test sets with a ratio of 3:1:1, which is consistent with existing ER studies~\cite{tu2022domain, ditto, deepmatcher}.

%

\stitle{Evaluation Metrics.}
In this paper, we evaluate the performance of ER approaches on both \emph{Accuracy} and \emph{Cost}.

\vspace{1mm} \noindent
(1) \textbf{Matching Accuracy.} Following existing ER studies~\cite{deepmatcher, ditto, robem, jointbert}, we use F1 score to measure the matching accuracy of an ER approach. 
Specifically, let ${\tt TP}$, ${\tt FP}$, ${\tt FN}$ denote the number of true positives (\ie matching pairs correctly identified), false positives (non-matching pairs incorrectly identified) and false negatives (matching-pairs incorrectly omitted) respectively. Then, we can respectively compute Precision and Recall as ${\tt P}={\tt TP}/({\tt TP}+{\tt FP})$ and ${\tt R}={\tt TP}/({\tt TP}+{\tt FN})$, and derive F1 score as harmonic mean of Precision and Recall, \ie ${\tt F1}=2\cdot {\tt P}\cdot {\tt R}/({\tt P} + {\tt R})$.

%

\vspace{1mm} \noindent
{(2) \textbf{Monetary Cost.}}
We evaluate an approach by considering its incurred monetary cost, which consists of two parts.
\bi
	\item \textbf{API Cost} measures how much an approach pays for calling the API of a proprietary LLMs (\eg GPT-3.5 and GPT-4). In particular, the API is priced per token. For example, according to the pricing of GPT API services\footnote{https://openai.com/pricing}, GPT-4 incurs $\$0.01$ / 1K tokens for input texts. 
	\item \textbf{Labeling Cost} measures how much an approach pays for labeling entity pairs to prepare \demos. To calculate the cost, we refer to the latest rates on the crowdsourcing platform, Amazon Mechanical Turk (AMT)~\footnote{https://www.mturk.com/} for text data labeling, which is $\$0.08$ per labeling task. Following the existing crowdsourcing approach to ER~\cite{DBLP:journals/pvldb/WangKFF12}, we group ten entity pairs into one labeling task and ask the crowd to label them in batch. Based on this, we estimate the cost of labeling one entity pair as $\$0.008$. 
\ei

%
%

\stitle{Baselines.}
We consider two types of baselines. The first type is the SOTA PLM-based approaches to ER, including Ditto~\cite{ditto}, JointBert~\cite{jointbert} and RobEM~\cite{robem}. The other type is the LLM-based approaches~\cite{DBLP:journals/pvldb/NarayanCOR22} to ER via in-context learning, equipped with manually designed prompts. We briefly describe the methods.

\vspace{1mm} \noindent
{(1) {\bf Ditto}~\cite{ditto}} is a well-recognized PLM-based approach to ER, which utilizes pre-trained language model RoBerta~\cite{liu2019roberta} and employs labeled entity pairs for fine-tuning. We use the code and default setting of Ditto in its original paper~\cite{ditto}.

\vspace{1mm} \noindent
{(2) {\bf JointBert}~\cite{jointbert}} is a dual-objective training method for BERT that combines binary matching and multi-class classification for entity matching. We use the code provided from~\cite{codeofjointbert}. We select the uncased base versions of BERT for JointBert and set all the hyper-parameters as default as in the original paper.

\vspace{1mm} \noindent
{(3) {\bf RobEM}~\cite{robem}} is a recent work that investigates the robustness of PLM-based ER methods with varying data distributions and identifies data imbalance as a critical issue. To solve this, it proposes simple yet effective modifications to enhance PLMs and achieves superior performance on ER. We run its original code from~\cite{codeofrobem} and keep all the setting as default.

\vspace{1mm} \noindent
{(4) {\bf ManualPrompt}~\cite{DBLP:journals/pvldb/NarayanCOR22}} is a pioneering initiative that uses LLMs (GPT-3) for ER as well as other data wrangling tasks. Similar to our work, it also employs in-context learning to answer ER questions. However, the key difference is that ManualPrompt utilizes standard prompting (\ie asking \pairs one by one) and manually designed \demos. We reproduce the results of ManualPrompt by using the prompts published by its original paper~\cite{DBLP:journals/pvldb/NarayanCOR22}.

\noindent

\stitle{Implementation Details.}
We briefly present the implementation details of our proposed framework as follows.

\vspace{1mm} \noindent
{(1) {\bf Batch Prompting}.}
We implement the design choices in Table~\ref{tbl:design_space} for question batching and demonstration selection, and compare them on both matching accuracy and monetary cost.
For question batching, we set the batch size to 8, which ensures that none of the design choices exceeds the maximum token limit of LLMs' text input, and employ the DBSCAN algorithm~\cite{DBLP:conf/kdd/EsterKSX96} for question clustering. 
For fair comparison of demonstration selection strategies (\ie fixed, Top$k$-\at{batch} and Top$k$-\at{question}), we choose $8$ \demos for each batch. For our covering-based strategy, we calculate the threshold $t$ by first computing the distances between all \pairs and then taking the $8$-th percentile as $t$ since $8$-th percentile can achieve great balance between cost and accuracy: with smaller $t$, the labeling cost will become larger while larger $t$ will degrade the matching accuracy.

\vspace{1mm} \noindent
{(2) {\bf Large Language Models}.}
In our experiments, we use GPT-3.5-turbo-0301, or GPT-3.5-03 for short, as the default LLM, where 0301 means that the model version was finalized on March 1st. In particular, according to the guideline of OpenAI\footnote{https://platform.openai.com/docs/api-reference/completions}, we set the temperature parameter of GPT-3.5-03 as 0.01. 
Moreover, we also investigate other proprietary LLMs, GPT-3.5-turbo-0613 (or GPT-3.5-06 for short) and GPT-4-1106-preview (or GPT-4 for short), as well as a very recent open-source LLM, LLama2-chat-70B~\cite{touvron2023llama}.

\subsection{Comparing Batch Prompting with Standard Prompting}
\label{subsec:exp-batch-vs-standard}

\noindent
\textbf{Exp-1: How does \emph{Batch Prompting} compare with \emph{Standard Prompting}?} 
We conduct experiments to compare batch prompting with standard prompting on matching accuracy and monetary cost.
For fair comparison, we use the same $8$ fixed \demos, which are selected randomly, for both approaches. In this case, we only need to consider the API cost, as labeling costs of both approaches are the same. Moreover, we run the experiments for three times, and compute mean and standard variance of the obtained F1 scores.

The experimental results are reported in Table~\ref{tbl:bp_vs_sp}.
We can see that, batch prompting significantly outperforms standard prompting on both accuracy and cost.
First, batch prompting improves F1 score by \textbf{1.3\%-30.6\%} on all datasets except Beer. The reason that batch prompting performs worse than standard prompting on the Beer dataset is that the dataset is very small (with only $91$ pairs for testing), and the two methods actually output very similar matching results. Moreover, we can also observe that batch prompting is more stable than standard prompting, \ie achieving much smaller standard variance.
Second, compared with standard prompting, batch prompting can achieve \textbf{4x-7x} cost saving on API callings.

\begin{table}[t!]
    \centering
    \caption{\bf{Comparing Batching Promting with Standard Prompting on Matching Accuracy and API Cost (The best results are bolded). }}
    \renewcommand{\arraystretch}{1.12}
    \vspace{-1mm}
    \begin{tabular}{|c|c||c|c|}
        \hline
        \textbf{Dataset}               & ~\textbf{Metric}~ & ~\textbf{Standard Prompting}~ & ~\textbf{Batch Prompting}~ \\ \hline \hline
        \multirow{2}{*}{\textbf{WA}}   & \textbf{F1}       & $67.54_{\pm 8.08}$            & $\bm{78.92}_{\pm 0.32}$    \\ \cline{2-4}
                                       & \textbf{API (\$)} & $1.43$                        & $\bm{0.33}$                \\ \hline \hline
        \multirow{2}{*}{\textbf{AB}}   & \textbf{F1}       & $65.70_{\pm 10.81}$           & $\bm{85.79}_{\pm 1.01}$    \\ \cline{2-4}
                                       & \textbf{API (\$)} & $1.10$                        & $\bm{0.24}$                \\ \hline \hline
        \multirow{2}{*}{\textbf{AG}}   & \textbf{F1}       & $53.72_{\pm 3.88}$            & $\bm{61.07}_{\pm 0.83}$    \\ \cline{2-4}
                                       & \textbf{API (\$)} & $1.29$                        & $\bm{0.29}$                \\ \hline \hline
        \multirow{2}{*}{\textbf{DS}}   & \textbf{F1}       & $75.08_{\pm 6.03}$            & $\bm{80.79}_{\pm 1.72}$    \\ \cline{2-4}
                                       & \textbf{API (\$)} & $5.31$                        & $\bm{1.22}$                \\ \hline \hline
        \multirow{2}{*}{\textbf{DA}}   & \textbf{F1}       & $85.96_{\pm 4.45}$            & $\bm{92.10}_{\pm 0.88}$    \\ \cline{2-4}
                                       & \textbf{API (\$)} & $2.93$                        & $\bm{0.63}$                \\ \hline \hline
        \multirow{2}{*}{\textbf{FZ}}   & \textbf{F1}       & $89.95_{\pm 3.67}$            & $\bm{94.13}_{\pm 1.11}$    \\ \cline{2-4}
                                       & \textbf{API (\$)} & $0.19$                        & $\bm{0.04}$                \\ \hline \hline
        \multirow{2}{*}{\textbf{IA}}   & \textbf{F1}       & $90.59_{\pm 0.94}$            & $\bm{91.75}_{\pm 0.84}$    \\ \cline{2-4}
                                       & \textbf{API (\$)} & $0.06$                        & $\bm{0.01}$                \\ \hline \hline
        \multirow{2}{*}{\textbf{Beer}} & \textbf{F1}       & $\bm{91.11}_{\pm 2.22}$       & $88.31_{\pm 2.60}$         \\ \cline{2-4}
                                       & \textbf{API (\$)} & $0.07$                        & $\bm{0.01}$                \\ \hline
    \end{tabular}
    \label{tbl:bp_vs_sp}
    \vspace{-1em}
\end{table}

\begin{figure}[!t]
    \centering
    \begin{subfigure}{0.492\columnwidth}
        \includegraphics[width=\textwidth]{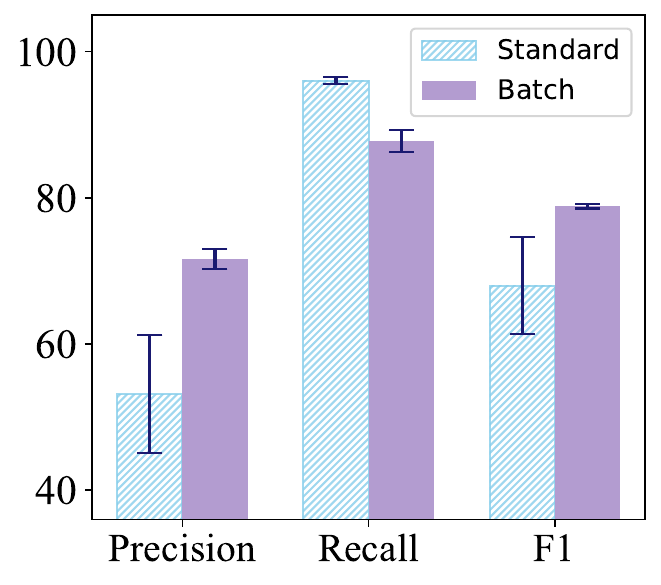}
        \caption{The WA Dataset}
        \label{fig:bp_vs_sp_details_wa}
    \end{subfigure}
    \hspace{-1mm}
    \begin{subfigure}{0.492\columnwidth}
        \includegraphics[width=\textwidth]{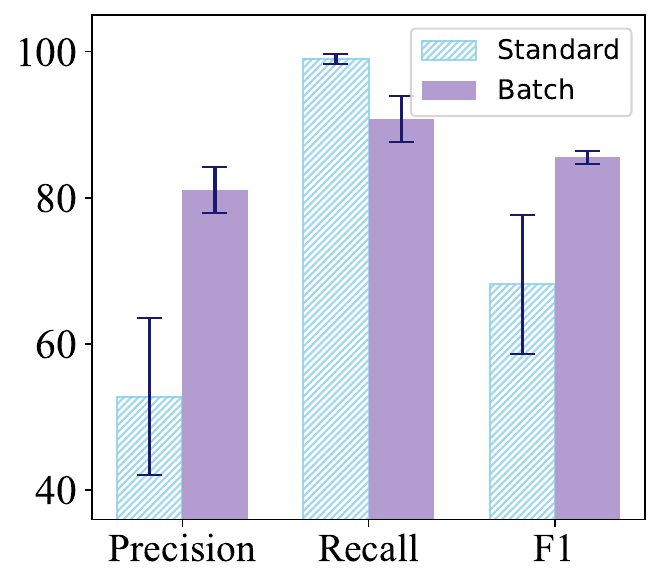}
        \caption{The AB Dataset}
        \label{fig:bp_vs_sp_details_ab}
    \end{subfigure}
    \caption{Comparing Batch Prompting and Standard Prompting on Recall, Precision and F1, where the two methods are denoted as ``Standard'' and ``Batch'' respectively.}
    \label{fig:bp_vs_sp_details}
    \vspace{-1em}
\end{figure}

While it is intuitive that batch prompting can save cost, it is somewhat surprising that it can also significantly improve the accuracy.
Thus, we conduct a detailed analysis to report Precision and Recall on WA and AB datasets, as shown in {Figure~\ref{fig:bp_vs_sp_details}}.
We can see batch prompting achieves much higher Precision than standard prompting, while their Recall scores are comparable. This is mainly attributed to the batching mechanism, where the LLM can refer to not only the provided \demos, but also the answers generated for previous \pairs within the same batch. 
This may help the LLM to identify some key characteristics that are useful to differentiate the entities.
For example, on the WA dataset, batch prompting can help the LLM to focus on a critical attribute ``$\term{modelno}$'', and enable the LLM to understand entities with different ``$\term{modelno}$'' tend to be non-matching pairs.

\vspace{1mm}
\noindent
\textbf{Finding 1: Batch prompting can not only bring {4x-7x} cost saving, but also achieve higher and more stable matching accuracy than standard prompting.}

\subsection{Exploring Design Space of Batch Prompting for ER}
\label{subsec:exp-design-space}

\begin{table*}[t!]
    \centering
   \vspace{-1em}
    \caption{\bf{Exploring the Design Space of Three Question Batching Methods and Four Demonstration Selection Methods (The best results are bolded and the second best results are underlined).}
    }
    \renewcommand{\arraystretch}{1.12}
    \tabcolsep=0.08cm
    \begin{tabular}{|c|c||c|c|c|>{\columncolor{gray!25}}c|c|c|c|>{\columncolor{gray!25}}c|c|c|c|>{\columncolor{gray!25}}c|}
        \hline
        \multirow{2}{*}{\textbf{Dataset}} & \multirow{2}{*}{\textbf{Metric}} & \multicolumn{4}{c|}{\textbf{Random Question Batching}} & \multicolumn{4}{c|}{\textbf{Similarity-based Question Batching}} & \multicolumn{4}{c|}{\textbf{Diversity-based Question Batching}}                                                                                                                                                                               \\ \cline{3-14}
                                          &                                  & \textbf{Fix}                               & \textbf{Top$k$-\at{batch}}                        & \textbf{Top$k$-\at{question}}                         & \textbf{Cover} & \textbf{Fix}  & \textbf{Top$k$-\at{batch}} & \textbf{Top$k$-\at{question}} & \textbf{Cover} & \textbf{Fix}  & \textbf{Top$k$-\at{batch}} & \textbf{Top$k$-\at{question}} & \textbf{Cover}  \\
        \hline \hline
        \multirow{3}{*}{\textbf{WA}}      & \textbf{F1}                      & 78.92                                      & 79.15                                       & 79.06                                       & 78.64          & 73.50         & 77.43                & 78.30               & 76.43          & 79.24         & 78.87                & \uline{80.18}       & \textbf{80.66}  \\ \cline{2-14} 
                                          & \textbf{API (\$)}                & 0.33                                       & 0.34                                        & 0.35                                        & 0.30           & 0.34          & 0.34                 & 0.35                & \textbf{0.24}  & 0.35          & 0.34                 & 0.34                & \uline{0.28}    \\ \cline{2-14} 
                                          & \textbf{Label (\$)}              & \textbf{0.06}                              & 11.53                                       & 12.63                                       & \uline{0.34}   & \textbf{0.06} & 14.15                & 12.63               & \uline{0.34}   & \textbf{0.06} & 13.30                & 12.63               & \uline{0.34}    \\ 
        \hline \hline
        \multirow{3}{*}{\textbf{AB}}      & \textbf{F1}                      & 85.79                                      & 86.24                                       & 86.79                                       & 85.71          & 85.19         & 85.65                & 87.02               & 87.16          & 85.03         & 86.38                & \uline{87.91}       & \textbf{88.38}  \\ \cline{2-14} 
                                          & \textbf{API (\$)}                & 0.24                                       & 0.23                                        & 0.24                                        & 0.21           & 0.24          & 0.23                 & 0.24                & \textbf{0.20}  & 0.24          & 0.23                 & 0.24                & \uline{0.20}    \\ \cline{2-14} 
                                          & \textbf{Label (\$)}              & \textbf{0.06}                              & 10.86                                       & 6.07                                        & \uline{0.28}   & \textbf{0.06} & 10.86                & 6.07                & \uline{0.28}   & \textbf{0.06} & 11.21                & 6.07                & \uline{0.28}    \\  
        \hline \hline
        \multirow{3}{*}{\textbf{AG}}      & \textbf{F1}                      & 61.07                                      & 61.82                                       & 61.90                                       & 60.69          & 58.90         & 60.74                & 60.96               & 60.62          & 60.24         & 57.85                & \textbf{64.57}      & \uline{62.16}   \\  \cline{2-14} 
                                          & \textbf{API (\$)}                & 0.29                                       & 0.30                                        & 0.30                                        & 0.25           & 0.30          & 0.30                 & 0.30                & \textbf{0.25}  & 0.29          & 0.30                 & 0.30                & \uline{0.25}    \\ \cline{2-14}
                                          & \textbf{Label (\$)}              & \textbf{0.06}                              & 14.20                                       & 9.70                                        & \uline{0.23}   & \textbf{0.06} & 14.09                & 9.70                & \uline{0.23}   & \textbf{0.06} & 13.84                & 9.69                & \uline{0.23}    \\ 
        \hline \hline
        \multirow{3}{*}{\textbf{DS}}      & \textbf{F1}                      & 80.79                                      & 82.49                                       & \uline{83.55}                                       & 82.36          & 76.44         & 73.78                & 77.09               & 75.59          & 79.07         & 79.80                & 83.46       & \textbf{83.70}  \\ \cline{2-14} 
                                          & \textbf{API (\$)}                & 1.22                                       & 1.27                                        & 1.28                                        & 1.13           & 1.31          & 1.27                 & 1.29                & \textbf{1.04}  & 1.27          & 1.15                 & 1.28                & \uline{1.12}    \\ \cline{2-14} 
                                          & \textbf{Label (\$)}              & \textbf{0.06}                              & 35.38                                       & 27.94                                       & \uline{0.31}   & \textbf{0.06} & 35.92                & 28.24               & \uline{0.31}   & \textbf{0.06} & 35.96                & 28.24               & \uline{0.31}    \\  
        \hline \hline
        \multirow{3}{*}{\textbf{DA}}      & \textbf{F1}                      & 92.10                                      & 93.00                                       & 93.62                               		  & 92.32          & 91.59         & 92.42                & 92.44               & 92.06          & 92.27         & 94.21                & \uline{94.28}      & \textbf{94.96}           \\ \cline{2-14} 
                                          & \textbf{API (\$)}                & 0.63                                       & 0.62                                        & 0.63                                        & 0.54   		   & 0.62          & 0.62                 & 0.63                & \textbf{0.50}  & 0.62          & 0.62                 & 0.63                & \uline{0.53}            \\  \cline{2-14} 
                                          & \textbf{Label (\$)}              & \textbf{0.06}                              & 15.50                                       & 14.61                                       & \uline{0.32}   & \textbf{0.06} & 15.50                & 14.61               & \uline{0.32}   & \textbf{0.06} & 15.09                & 14.61               & \uline{0.32}    \\  [1pt]
        \hline \hline
        \multirow{3}{*}{\textbf{FZ}}      & \textbf{F1}                      & 94.13                                      & 93.33                                       & 95.24                                       & 93.33          & 95.24         & 90.48                & 93.02               & 92.68          & 93.02         & 88.37                & \uline{95.24}       & \textbf{100.00} \\ \cline{2-14} 
                                          & \textbf{API (\$)}                & 0.04                                       & 0.04                                        & 0.03                                        & \uline{0.03}   & 0.04          & 0.04                 & 0.04                & \textbf{0.03}  & 0.04          & 0.04                 & 0.04                & 0.03            \\ \cline{2-14}
                                          & \textbf{Label (\$)}              & \textbf{0.06}                              & 1.18                                        & 1.27                                        & \uline{0.30}   & \textbf{0.06} & 1.25                 & 1.32                & \uline{0.30}   & \textbf{0.06} & 1.18                 & 1.27                & \uline{0.30}    \\ 
        \hline \hline
        \multirow{3}{*}{\textbf{IA}}      & \textbf{F1}                      & 91.75                                      & 94.74                                       & 94.55                                       & 92.59          & 92.59         & 94.34                & 96.30               & 92.86          & 88.00         & 94.55                & \textbf{98.17}      & \uline{96.43}   \\ \cline{2-14} 
                                          & \textbf{API (\$)}                & 0.01                                       & 0.01                                        & 0.01                                        & \uline{0.01}   & 0.01          & 0.01                 & 0.01                & \textbf{0.01}  & 0.01          & 0.01                 & 0.01                & 0.01            \\ \cline{2-14} 
                                          & \textbf{Label (\$)}              & \textbf{0.06}                              & 0.60                                        & 0.56                                        & \uline{0.16}   & \textbf{0.06} & 0.69                 & 0.56                & \uline{0.16}   & \textbf{0.06} & 0.42                 & 0.56                & \uline{0.16}    \\ 
        \hline \hline
        \multirow{3}{*}{\textbf{Beer}}    & \textbf{F1}                      & 88.31                                      & 76.92                                       & 81.48                                       & 89.66          & 85.71         & 84.62                & 81.48               & 88.89          & \uline{92.86} & 89.66                & 89.66               & \textbf{96.55}  \\  \cline{2-14} 
                                          & \textbf{API (\$)}                & 0.01                                       & 0.01                                        & 0.01                                        & \uline{0.01}   & 0.01          & 0.01                 & 0.01                & \textbf{0.01}  & 0.01          & 0.01                 & 0.01                & 0.01            \\ \cline{2-14} 
                                          & \textbf{Label (\$)}              & \textbf{0.06}                              & 0.65                                        & 0.66                                        & \uline{0.14}   & \textbf{0.06} & 0.68                 & 0.66                & \uline{0.14}   & \textbf{0.06} & 0.64                 & 0.62                & \uline{0.14}    \\ 
        \hline
    \end{tabular}
    \label{tbl:transposed_main_results}
\end{table*}

\noindent
\textbf{Exp-2: What are effective strategies in our design space of question batching and demonstration selection?}
We explore the design space shown in Table~\ref{tbl:design_space} by comparing the 12 combinations of three question batching methods and four demonstration selection methods.
From the experimental results reported in Table~\ref{tbl:transposed_main_results}, we have the following observations.

\stitle{Evaluation on question batching.}
As reported in Table~\ref{tbl:transposed_main_results}, the diversity-based question batching achieves the highest overall F1 scores.
Moreover, it is interesting to see that the similarity-based question batching performs the worst on matching accuracy, even achieving lower F1 scores than the random question batching. This is because the \pairs within a batch is very similar, thus making the LLM difficult to differentiate entities by comparing different \pairs. Consequently, the LLM tends to produce identical answers for various questions, leading to degradation of matching accuracy.  
On the other hand, we can see that different question batching strategies have similar results on API cost and labeling cost, given varying demonstration selection methods. The reason is straightforward since prompts of different question batching strategies have similar amounts of tokens.

\stitle{Evaluation on demonstration selection.}
Observing Table~\ref{tbl:transposed_main_results} again, we can see that Top$k$-\at{question} and our covering-based strategy (denoted as Cover) outperform other strategies on accuracy, while the F1 scores of these two strategies are comparable. For example, under diversity-based batching, Top$k$-\at{question} yields the highest F1 score on $2$ datasets, while Cover is the best on the remaining $6$ datasets.
This is because both Top$k$-\at{question} and Cover aim to select relevant \demos for each individual \pair within a batch, which is helpful for the LLM to understand varying cases of ER.

On the other hand, Cover is much more cost-effective than Top$k$-\at{question} on demonstration labeling, \eg brings 10x-100x labeling cost savings on the former five large datasets and 5x savings on the latter three small datasets. The results validate the effectiveness of our \emph{covering-based} mechanism: by selecting a minimal set of \demos that cover all \pairs in a batch, we can significantly reduce the number of required \demos, and thus save the labeling cost.

\vspace{1mm}
\noindent
\textbf{Finding 2: The design choice that combines Diversity-based Question Batching and our Covering-based Demonstration Selection is the most favorable, \ie achieving the highest accuracy while incurring the lowest cost.}

\subsection{Comparing with PLM-based Approaches to ER}
\label{subsec:exp-compare-plms}
\begin{figure*}[!t]
  \centering
  
  \begin{subfigure}{0.24\textwidth}
    \includegraphics[width=\textwidth]{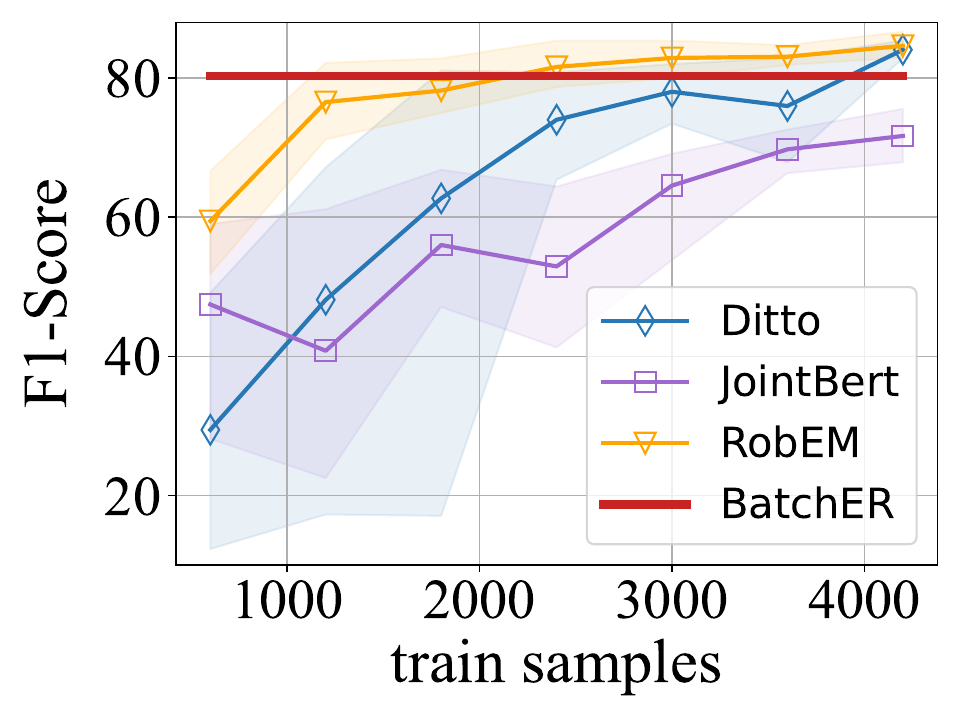}
    \caption{WA}
    \label{fig:compare_with_plms_em_wa}
  \end{subfigure}
  \hfill
  \begin{subfigure}{0.24\textwidth}
    \includegraphics[width=\textwidth]{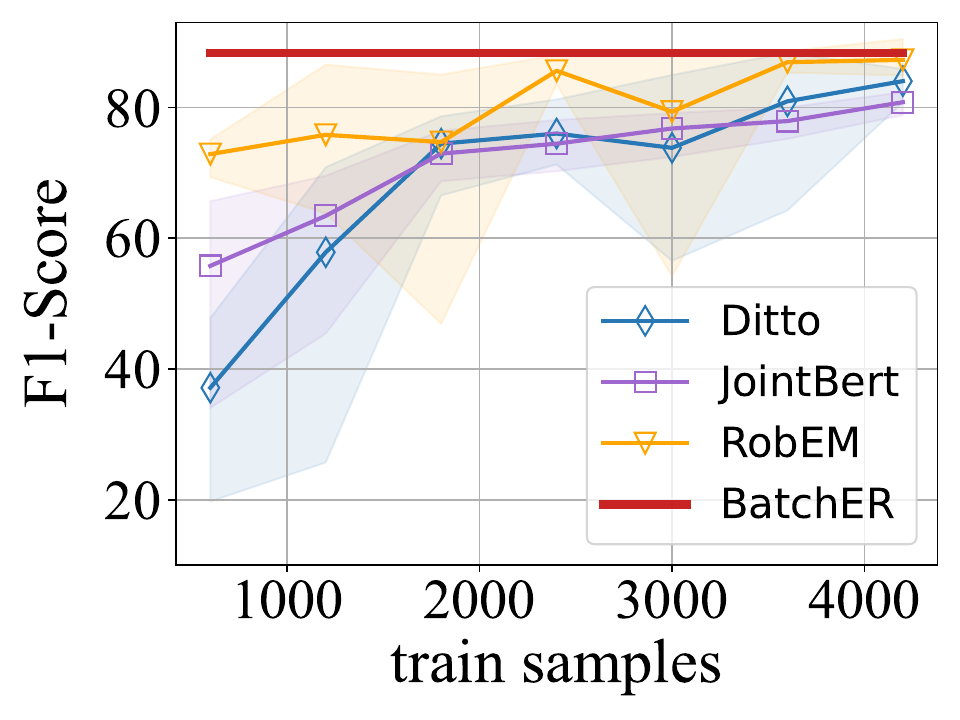}
    \caption{AB}
    \label{fig:compare_with_plms_abt_buy}
  \end{subfigure}
  \hfill
  \begin{subfigure}{0.24\textwidth}
    \includegraphics[width=\textwidth]{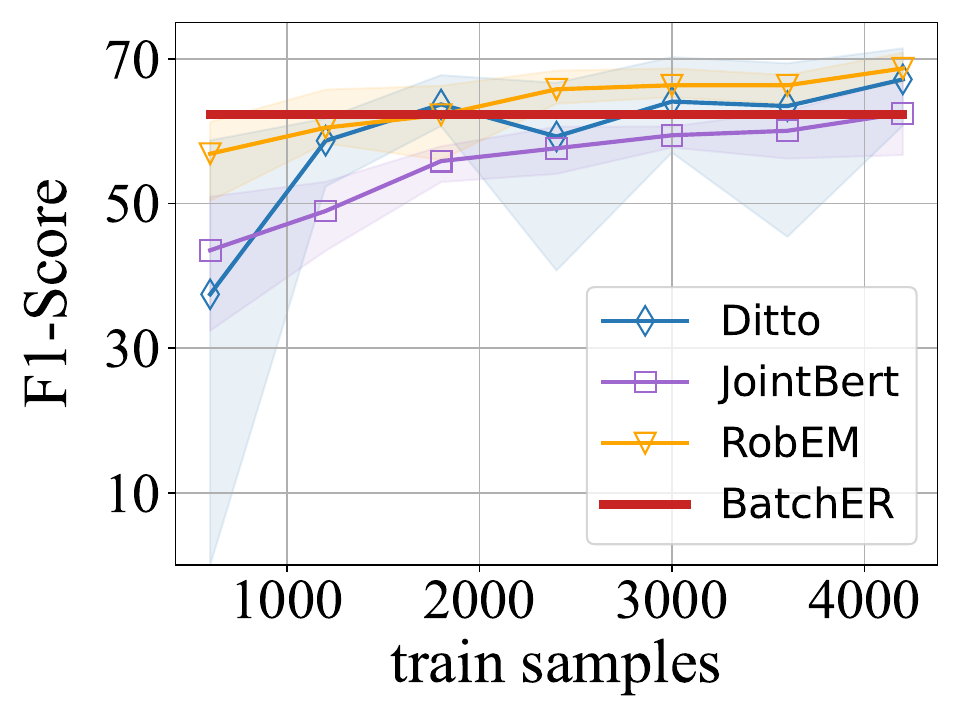}
    \caption{AG}
    \label{fig:compare_with_plms_em_ag}
  \end{subfigure}
  \hfill
  \begin{subfigure}{0.24\textwidth}
    \includegraphics[width=\textwidth]{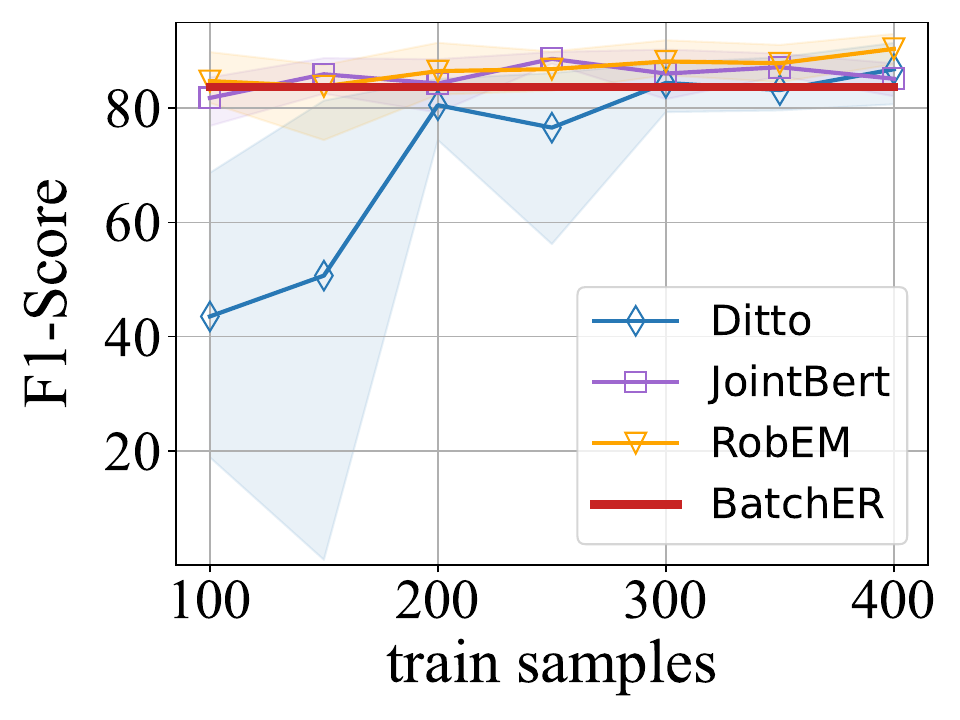}
    \caption{DS}
    \label{fig:compare_with_plms_em_ds}
  \end{subfigure}
  
  \vspace{1ex}
  
  \begin{subfigure}{0.24\textwidth}
    \includegraphics[width=\textwidth]{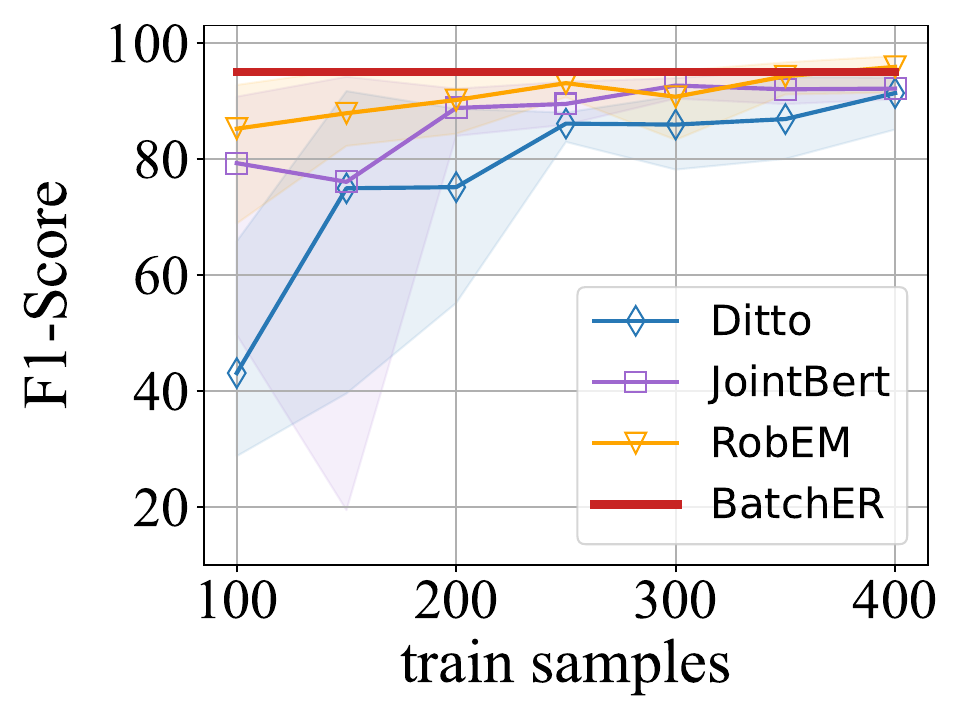}
    \caption{DA}
    \label{fig:compare_with_plms_em_da_dirty}
  \end{subfigure}
  \hfill
  \begin{subfigure}{0.24\textwidth}
    \includegraphics[width=\textwidth]{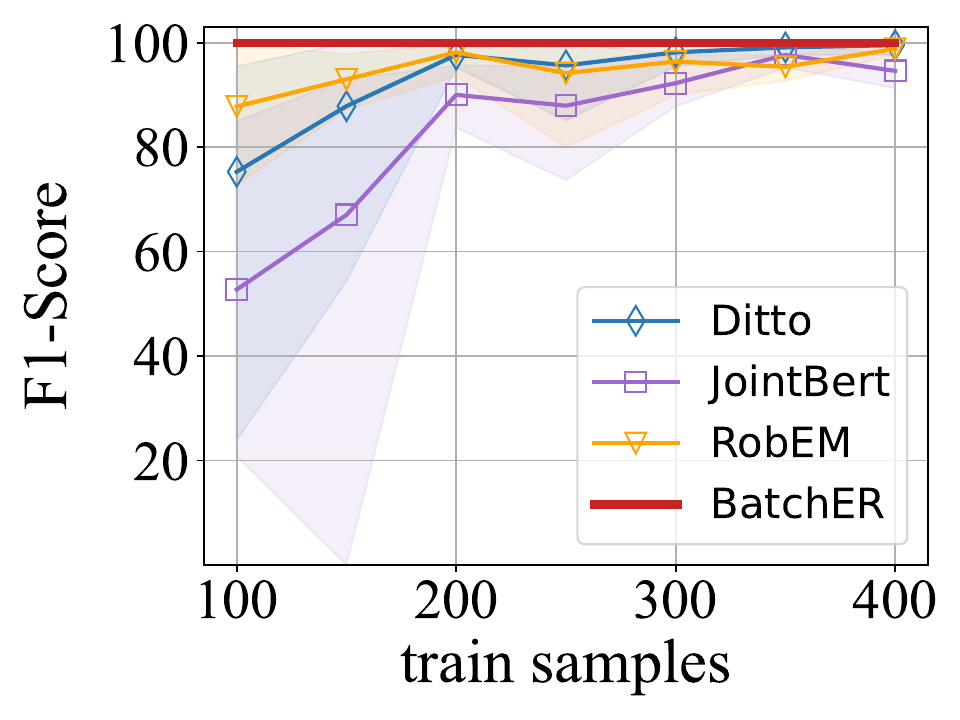}
    \caption{FZ}
    \label{fig:compare_with_plms_em_fz}
  \end{subfigure}
  \hfill
  \begin{subfigure}{0.24\textwidth}
    \includegraphics[width=\textwidth]{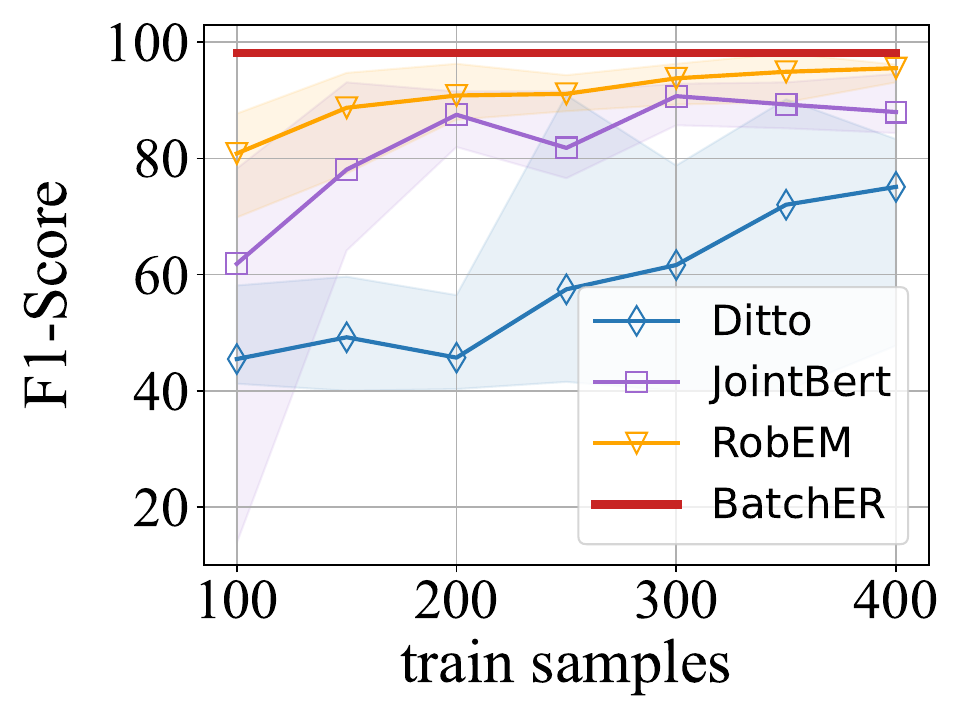}
    \caption{IA}
    \label{fig:compare_with_plms_em_ia}
  \end{subfigure}
  \hfill
  \begin{subfigure}{0.24\textwidth}
    \includegraphics[width=\textwidth]{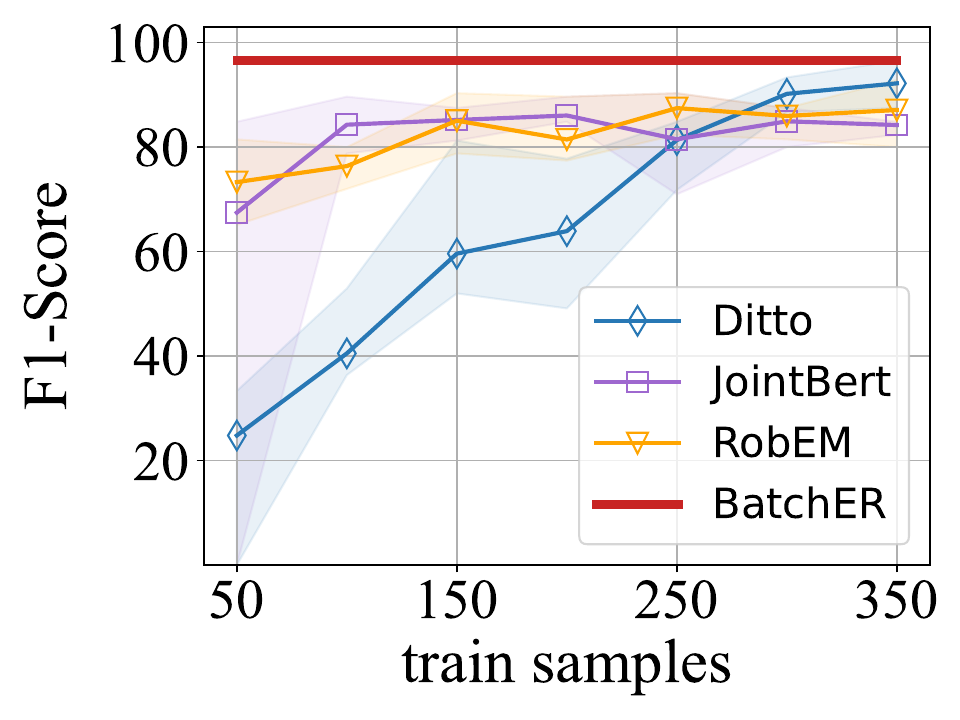}
    \caption{Beer}
    \label{fig:compare_with_plms_em_beer}
  \end{subfigure}  
  \caption{Comparing our Batching Prompting framework \sys with existing PLM-based approaches to ER.}
  \label{fig:compare_with_plms}
     \vspace{-1em}
\end{figure*}

\noindent
\textbf{Exp-3: How does our \sys framework compare with PLM-based approaches to ER?}
We compare our framework with the PLM-based approaches mentioned in Section~\ref{subsec:exp-setup}, by varying the size of training set for these approaches.
Note that we use the best design choices shown in Table~\ref{tbl:transposed_main_results}, \ie Diversity-based Question Batching and Covering-based Demonstration Selection, as the default setting.

Figure~\ref{fig:compare_with_plms} shows the experimental results on the eight datasets, where the results of our framework are represented as red solid lines.
Not surprisingly, our framework is \emph{much more cost-effective} than Ditto~\cite{ditto}, JointBert~\cite{jointbert} and RobEM~\cite{robem}. For example, on the WA, AB and AG datasets, the three PLM-based methods require at least 2000 training samples to achieve a similar F1 score of our framework. In contrast, our framework requires no more than 50 labeled samples on all the datasets. According to our cost calculation method in Section~\ref{subsec:exp-setup}, the monetary cost incurred by these PLM-based approaches is about \textbf{300x-400x} larger than our overall cost (\ie API cost plus labeling cost).
Furthermore, we also observe that once models like RobEM catching up with the F1 score of our framework, additional training samples do not substantially increase the performance; on some datasets (\eg FA, IA and Beer), even the entire training set is insufficient for the baselines to reach the F1 score of our framework. 

\vspace{1mm}
\noindent
\textbf{Finding 3: With much less labeled data, our batch prompting framework achieves competitive performance with PLM-based method trained with hundreds of or even thousands of labeled matching/non-matching entity pairs.}
\subsection{Comparing with Manual Prompting for ER}
\label{subsec:exp-compare-llms}

\begin{table}[t!]
    \centering
    \caption{\bf{Comparing Batching Prompting with Manual Prompting (The best results are bolded).
    }}

    \renewcommand{\arraystretch}{1.12}
    \vspace{-1mm}
    \begin{tabular}{|c|c||c|c|}
        \hline
        \textbf{Dataset}                   & ~\textbf{Metric}~ & ~\textbf{Manual Prompting}~ & ~\textbf{Batch Prompting}~ \\ \hline \hline
        \multirow{2}{*}{\textbf{WA}}       & \textbf{F1}       & \textbf{82.63}              & 80.66                      \\ \cline{2-4}
                                           & \textbf{API (\$)} & 1.40                        & \textbf{0.28} \\
        \hline \hline
        \multirow{2}{*}{\textbf{AG}}       & \textbf{F1}       & \textbf{65.40}              & 62.16                      \\ \cline{2-4}
                                           & \textbf{API (\$)} & 1.65                        & \textbf{0.25}   
                                           \\ 
        \hline \hline
        \multirow{2}{*}{\textbf{DS}}       & \textbf{F1}       & 70.44                       & \textbf{83.70}             \\ \cline{2-4}
                                           & \textbf{API (\$)} & 5.87                        & \textbf{1.12}              \\
        \hline \hline
        \multirow{2}{*}{\textbf{DA}}       & \textbf{F1}       & 94.90                       & \textbf{94.96}             \\ \cline{2-4}
                                           & \textbf{API (\$)} & 2.65                        & \textbf{0.53}              \\ 
        \hline \hline
        \multirow{2}{*}{\textbf{FZ}}       & \textbf{F1}       & 97.67                       & \textbf{100}               \\ \cline{2-4}
                                           & \textbf{API (\$)} & 0.14               & \textbf{0.03}                       \\ 
        \hline \hline
        \multirow{2}{*}{\textbf{IA}}       & \textbf{F1}       & \textbf{98.11}              & 96.43                      \\ \cline{2-4}
                                           & \textbf{API (\$)} & 0.05            & \textbf{0.01}                       \\
        \hline \hline
        \multirow{2}{*}{\textbf{Beer}}     & \textbf{F1}       & 92.23                       & \textbf{96.55}             \\ \cline{2-4}
                                           & \textbf{API (\$)} & 0.05               & \textbf{0.01}                       \\ 
        \hline
    \end{tabular}
    \label{tbl:bp_vs_mp}
    \vspace{-1em}
\end{table}

\noindent
\textbf{Exp-4: How does our \sys framework compare with LLM-based approaches to ER?}
We compare our framework with the existing LLM-based approach~\cite{DBLP:journals/pvldb/NarayanCOR22}, equipped with manually designed prompts, including hand-picked \demos. The results are reported in Table~\ref{tbl:bp_vs_mp}. The reason for the absence of a comparison for the Abt-Buy dataset in the Table~\ref{tbl:bp_vs_mp} is that ManualPrompt approach~\cite{DBLP:journals/pvldb/NarayanCOR22} is not tested on this dataset. We can see that, with only 20\% of the API cost, our batch prompting framework can achieve comparable F1 score, compared with the ManualPrompt approach. In particular, on four datasets (DS, DA, FZ, Beer), our framework even outperforms ManualPrompt. The results implies that batch prompting, despite requiring cost of labeling selected \demos, may still be more practical than ManualPrompt, which requires domain experts for prompt designing.

\vspace{1mm}
\noindent
\textbf{Finding 4: Our automatic batch prompting framework achieves comparable or even better F1 scores with manual prompting methods for LLMs, with much less API cost.}
\subsection{Evaluation on Different Underlying LLMs}
\label{subsec:exp-given-llms}

\begin{table}[t!]
    \centering
    \caption{\bf{Evaluating Different Underlying LLMs on Matching Accuracy and API Cost (The best results are bolded and the second best results are underlined).}}
    \renewcommand{\arraystretch}{1.12}
    \vspace{-1mm}
    \begin{tabular}{|c|c||c|c|c|}
        \hline
        \textbf{Dataset}                   & ~\textbf{Metric}~ & ~\textbf{GPT-3.5-03}~ & ~\textbf{GPT-3.5-06}~ & ~\textbf{GPT-4}~ \\ \hline \hline
        \multirow{2}{*}{\textbf{WA}}      & \textbf{F1}                       & \uline{80.66}                                      & 80.32                & \textbf{81.22}               \\ \cline{2-5}
                                          & \textbf{API (\$)}                  & \textbf{0.28}                                       & \textbf{0.28}                 & \uline{2.81}            \\ \cline{2-5}
        \hline \hline
        \multirow{2}{*}{\textbf{AB}}      & \textbf{F1}                       & \textbf{88.38}                                      & 69.08                & \uline{85.22}                \\ \cline{2-5}
                                          & \textbf{API (\$)}                  & \textbf{0.20}                                       & \textbf{0.20}                 & \uline{2.02}            \\ \cline{2-5}
        \hline \hline
        \multirow{2}{*}{\textbf{AG}}      & \textbf{F1}                       & \uline{62.16}                                      & 52.40                & \textbf{64.06}  \\ \cline{2-5}
                                          & \textbf{API (\$)}                  & \textbf{0.25}                                       & \textbf{0.25}                 & \uline{2.52}            \\ \cline{2-5}
        \hline \hline
        \multirow{2}{*}{\textbf{DS}}      & \textbf{F1}                       & \uline{83.70}                                      & 65.94                & \textbf{89.48}  \\ \cline{2-5}
                                          & \textbf{API (\$)}                  & \textbf{1.12}                                       & \textbf{1.12}                 & \uline{11.24}           \\ \cline{2-5}
        \hline \hline
        \multirow{2}{*}{\textbf{DA}}      & \textbf{F1}                       & \uline{94.96}                                      & 91.29                & \textbf{96.04}  \\ \cline{2-5}
                                          & \textbf{API (\$)}                  & \textbf{0.53}                                       & \textbf{0.53}                 & \uline{5.27}            \\ \cline{2-5}
        \hline \hline
        \multirow{2}{*}{\textbf{FZ}}      & \textbf{F1}                       & \textbf{100.00}                            & \uline{92.68}                & \textbf{100.00} \\ \cline{2-5}
                                          & \textbf{API (\$)}                  & \textbf{0.03}                                       & \textbf{0.03}                 & \uline{0.32}            \\ \cline{2-5}
        \hline \hline
        \multirow{2}{*}{\textbf{IA}}      & \textbf{F1}                       & \textbf{96.43}                             & 92.31                & \uline{94.34}           \\ \cline{2-5}
                                          & \textbf{API (\$)}                  & \textbf{0.01}                                       & \textbf{0.01}                 & \uline{0.09}            \\ \cline{2-5}
        \hline \hline
        \multirow{2}{*}{\textbf{Beer}}    & \textbf{F1}                       & \textbf{96.55}                             & 92.31                & \uline{96.30}           \\ \cline{2-5}
                                          & \textbf{API (\$)}                  & \textbf{0.01}                                       & \textbf{0.01}                 & \uline{0.11}            \\ \cline{2-5}
        \hline
    \end{tabular}
    \label{tbl:transposed_llm}
\end{table}

\noindent
\textbf{Exp-5: What is performance of our approaches given various underlying LLMs?} We evaluate the performance of \sys on various underlying LLMs, including two versions of GPT-3.5 and GPT-4, which are mentioned in Section~\ref{subsec:exp-setup}. Note that we also evaluate the well-known open-source LLM, \textsc{Llama2}~\cite{DBLP:journals/corr/abs-2307-09288}. However,  
%
we find that \textsc{Llama2} is not suitable for batch prompting: When prompted to answer multiple questions, \textsc{Llama2} fails to produce any output in most cases. Thus, we omit the results of \textsc{Llama2}.

The experimental results are shown in Table~\ref{tbl:transposed_llm}. First, considering matching accuracy, GPT-4 achieves the best results on five datasets, demonstrating its superior capability on text comprehension and task solving. Moreover, we also find GPT-3.5-03 is comparable to GPT-4. Specifically, GPT-3.5-03 achieves the second highest F1 overall and the largest F1 difference from GPT-4 is less than 6.4\%. Second, as per the latest pricing, the token pricing of GPT-4 is \textbf{10x} higher than GPT-3.5, leads to considerably high API costs. To summarize, the results show that GPT-3.5-03 achieves the best trade-off between matching accuracy and monetary cost, making it a more favorable choice for practical applications.

\vspace{1mm}
\noindent
\textbf{Finding 5: As the underlying LLM of \sys, GPT3.5-03 achieves the best trade-off between matching accuracy and monetary cost.}

\subsection{Evaluation on Different Feature Extractors}
\label{subsec:exp-given-feat-extract}

\begin{table}[t!]
    \centering
    \caption{\bf{Evaluating Different Feature Extractors on Matching Accuracy (The best results are bolded).}}
    \renewcommand{\arraystretch}{1.12}
    \vspace{-1mm}
    \begin{tabular}{|c||c|c|c|}
        \hline
        \multirow{2}{*}{\textbf{Dataset}} & \multicolumn{2}{c|}{\textbf{Structure-aware}} & \textbf{Semantics-based}                       \\
        \cline{2-4}
                                          & \textbf{\sys-LR}                           & \textbf{\sys-JAC}      & \textbf{\sys-SEM} \\
        \hline \hline
        \textbf{WA}                       & \textbf{80.66}                                & 78.05                    & \uline{78.66}               \\  \cline{2-4}
        \hline \hline
        \textbf{AB}                       & \textbf{88.38}                                & 84.23                    & \uline{87.06}               \\  \cline{2-4}
        \hline \hline
        \textbf{AG}                       & \textbf{62.16}                                & \uline{59.90}                    & 59.20               \\  \cline{2-4}
        \hline \hline
        \textbf{DS}                       & \textbf{83.70}                                & \uline{81.27}                    & 80.91               \\  \cline{2-4}
        \hline \hline
        \textbf{DA}                       & \textbf{94.96}                                & \uline{92.70 }                   & 90.36               \\  \cline{2-4}
        \hline \hline
        \textbf{FZ}                       & \textbf{100.00}                               & 93.62                    & \uline{95.24}               \\  \cline{2-4}
        \hline \hline
        \textbf{IA}                       & \textbf{96.43}                                & 90.57                    & \uline{90.91}               \\  \cline{2-4}
        \hline \hline
        \textbf{Beer}                     & \textbf{96.55}                                & 89.66                    & \uline{91.67}               \\  \cline{2-4}
        \hline
    \end{tabular}
    \label{tbl:transposed_sim}
    \vspace{-1em}
\end{table}

\noindent
\textbf{Exp-6: What is performance of our approaches given different feature extractors?}
We examine the performance of \sys using different Feature Extractors described in Section~\ref{subsec:feature_extractor}, namely \sys-LR, \sys-JAC, and \sys-SEM. The former two feature extractor use Structure-aware Feature Extractor based on Levenshtein Ratio (LR) and Jaccard Similarity (JAC). The latter uses Semantics-based Feature Extractor based on SBERT embedding. Since their monetary cost is close, we only compare these three variants on F1 scores on the eight datasets.

As shown in Table~\ref{tbl:transposed_sim}, \sys-LR achieves the best performance on all the datasets while \sys-JAC and \sys-SEM achieve comparative results. This results validates that stucture-aware feature extractor can better capture the relevance between entity pairs in the ER scenario. Moreover, compared with \sys-JAC, \sys-LR is more sensitivity to string order and its superior precision in quantifying the similarity between two strings. For instance, considering two strings ``listen'' and ``silent'', the similarity score calculated using LR is 0.5, whereas with JAC, it is 0.89. This clearly demonstrates the former is better effectiveness in quantifying the similarity between the two strings, thus is more effective to generate feature vectors for entity pairs.

\vspace{1mm}
\noindent
\textbf{Finding 6: The structure-aware feature extractor is preferred for measuring distances among entity pairs in ER.}

\section{Related Work} 
\label{sec:related_work} 

\noindent \textbf{PLM-based Methods for Entity Resolution.}
Entity resolution is a popular data integration task that has been widely studied for decades. 
With the rise of deep learning, some approaches~\cite{DBLP:journals/pvldb/EbraheemTJOT18} leverage pre-trained word embeddings to improve the ER performance. However, these methods mainly use the non-contextual embeddings without considering the downstream tasks. Therefore, recent studies~\cite{ditto, jointbert, tu2023unicorn, tu2022domain} have focused on using Transformer-based PLMs to produce contextualized embeddings based on fine-tuning over downstream tasks. To be specific, Ditto~\cite{ditto} regards ER as a sequence-pair classification problem via Transformer, where domain knowledge  is injected to further improve the performance. JointBERT~\cite{jointbert} adopts a dual-objective training  paradigm for BERT. Specifically, besides predicting matching/non-matching pairs, JointBERT also incorporates a multi-class classification task to predict the entity identifier for each entity description of a pair. DADER~\cite{tu2022domain} focuses on leveraging the domain adaptation technique: given a labeled source dataset, it trains an ER model for another target dataset by aligning features of both datasets based on PLMs. Based on PLMs, Unicorn~\cite{tu2023unicorn} focuses on building a unified framework for data matching tasks, including ER. Unicorn uses a unified encoder for any pair of data to be predicted, and a mixture-of-experts module to align the semantics of multiple tasks. Although the above PLMs-based approaches can achieve a relatively good performance, they need plenty of labeled pairs for supervision, which are often expensive to acquire.

\stitle{LLM-based Methods for Entity Resolution.} With the size of pre-training data and model parameters scales, large-scale language models (LLMs) have gained an emergent capability called In-Context Learning (ICL) to learn from a few \demos without explicit model update~\cite{DBLP:conf/nips/BrownMRSKDNSSAA20, chowdhery2022palm}. 
Recent studies~\cite{DBLP:journals/pvldb/NarayanCOR22,DBLP:journals/corr/abs-2310-11244,zhang2023large} have focused on utilizing LLMs to tackle ER with less labeled pairs for supervision. Narayan et al.~\cite{DBLP:journals/pvldb/NarayanCOR22} are among the first to explore the capability of GPT3~\cite{DBLP:conf/nips/BrownMRSKDNSSAA20} for ER with manually designed \demos, which achieves remarkable performance compared with PLM-based methods. Since manual \demos require professional prompting engineering knowledge, Peeters et al.~\cite{DBLP:journals/corr/abs-2310-11244} propose to select relevant \demos based on $\mathtt{KNN}$ retrieval algorithm, where Jaccard similarity is utilized to measure the relevance. Moreover, Zhang et al.~\cite{zhang2023large} consider batch prompting for ER, which employs a straightforward random batching strategy with manually designed demonstrations.
Although question batching and demonstration selection have been considered in existing studies, these studies mainly rely on domain experts or develop heuristics for these two problem, and have not explored the combination of different demonstration selection and batching strategies. Compared to them, we utilize the power of ICL and propose a comprehensive framework \sys. We explore a design space to evaluate the performance of different design choices, and propose a covering-based demonstration selection strategy that effectively balances the trade-off between accuracy and cost.

\stitle{In-Context Learning for Data Management.} LLMs are capable to capture rich linguistic patterns and generate coherent text~\cite{agrawal2022large, DBLP:conf/nips/BrownMRSKDNSSAA20, touvron2023llama}, which have shown great success in a wide range of NLP tasks ~\cite{wan2023gptre, agrawal2022context,li2023unified}.
ICL is an emergent capability of LLMs that enables the model to learn from few demonstrations without explicit gradient update~\cite{chowdhery2022palm}. Recently, researchers have studied to leverage ICL to solve data management tasks, such as data discovery~\cite{DBLP:journals/corr/abs-2306-09610}, data cleaning and integration~\cite{DBLP:journals/pvldb/NarayanCOR22}, and data labeling~\cite{DBLP:journals/corr/abs-2311-00739}, and also study how to batch questions and select demonstrations. BatchPrompt~\cite{cheng2023batch} proposes to group multiple questions into one batch and query LLMs to answer one batch in an interface. In addition, both relevance-based~\cite{DBLP:conf/acl-deelio/LiuSZDCC22, DBLP:conf/coling/LeeLC22} and diversity-based~\cite{DBLP:conf/acl/LevyBB23, DBLP:conf/iclr/SuKWSWX0OZS023} strategies are proposed for demonstration selection.
Compared with these studies, as far as we know, we are the first to develop the batch prompting technique tailored to the ER task, and design new methods, such as covering-based demonstration selection and structure-aware feature extraction, which are shown to be effective for ER.

\section{Conclusion}
In this paper we have introduced a cost-effective batch prompting framework \sys for entity resolution, and explored the effectiveness of \sys under different design choices. We also devised a covering-based \demo selection strategy that achieves effective balance between accuracy and cost. We have conducted extensive experiments to evaluate different combinations of the choices in the design space with insightful empirical findings, as summarized using the six findings in Section~\ref{sec:exp}. These findings imply that our \sys framework is very cost-effective for ER, compared with not only PLM-based methods fine-tuned with extensive labeled data, but also LLM-based methods with manually designed prompting. We also provided guidance for selecting appropriate design choices for batch prompting.

\newpage
\bibliographystyle{IEEEtran}
\bibliography{citations/ref} 

\balance

\end{document}